%% file: chord_arxiv.tex
\documentclass[11pt,twocolumn,a4paper]{extarticle}

\usepackage[numbers,sort&compress]{natbib}
\usepackage{proxima_report}

\usepackage[utf8]{inputenc}
\usepackage[T1]{fontenc}
\usepackage{hyperref}
\usepackage{xurl}
\usepackage{booktabs}
\usepackage{amsfonts}
\usepackage{amsmath}
\usepackage{nicefrac}
\usepackage{microtype}
\usepackage{xcolor}
\usepackage{graphicx}
\usepackage{multirow}
\usepackage{enumitem}
\usepackage{wrapfig}

\usepackage{float}
\usepackage{flushend}
\usepackage{standalone}
\usepackage{tikz}
\usetikzlibrary{arrows.meta,calc,positioning,decorations.markings}

\newcommand{\R}{\mathbb{R}}
\newcommand{\X}{\mathcal{X}}

\newcommand{\Z}{\mathcal{Z}}
\newcommand{\F}{\mathcal{F}}
\newcommand{\E}{\mathbb{E}}
\newcommand{\relu}[1]{\left[#1\right]_+}
\newcommand{\method}{\texttt{CHORD}}
\newcommand{\vmec}{\texttt{VMEC}~}
\newcommand{\desc}{\texttt{DESC}~}

\renewcommand{\S}{Sec.~}

\definecolor{violationred}{HTML}{E06666}
\definecolor{feasiblegreen}{HTML}{6AA84F}
\definecolor{chordgreen}{HTML}{6AA84F}
\definecolor{labelgray}{HTML}{595959}
\definecolor{cvaered}{HTML}{EA9999}
\newcommand{\sv}[2]{({#1/60.0}, {(540-#2)/60.0})}

\shorttitle{\method: Learning Latent Spaces for Constrained Stellarator Design}
\reportdate{September 2026}
\affiliations{Proxima Fusion, Munich, Germany.\\[2pt]
\textsuperscript{$\star$}\,\href{mailto:scadena@proximafusion.com}{\texttt{scadena@proximafusion.com}}}

\title{\method: Learning Latent Spaces for Constrained Stellarator Design}

\author{Santiago A. Cadena\textsuperscript{$\star$} \quad Andrea Merlo}

\begin{document}
\raggedbottom
\setcounter{dbltopnumber}{2}
\renewcommand{\dbltopfraction}{0.9}
\renewcommand{\textfraction}{0.05}

\twocolumn[%
\maketitle

\begin{abstract}
In high-dimensional, PDE-constrained design, interpolating between known feasible solutions can reveal trade-offs without repeating expensive optimization campaigns. In stellarator design for fusion energy, direct interpolation of plasma boundary parameters can violate physical constraints even when both endpoints are feasible. Learned latent representations can capture nonlinear relationships among design variables, but latent interpolation can still cross infeasible regions, and conditioning on target metrics can become inconsistent with the decoded design. We introduce \method{}, a conditional variational autoencoder that penalizes predicted constraint violations along decoded latent interpolants during training, paired with self-consistent inference that aligns conditioning with the decoded design's predicted metrics.
Geometric benchmarks show the benefit of regularizing latent interpolants. In stellarator interpolation between distinct geometry families, \method{} reaches up to $81\,\%$ simulator-validated mid-segment feasibility, compared with $50\,\%$ for an unregularized conditional VAE and $28\,\%$ for direct interpolation of boundary parameters. Ablations show that both ingredients are needed for this gain. Beyond interpolation, latent-space optimization finds simulator-feasible designs, while self-consistent sampling produces more diverse feasible alternatives.
On the ConStellaration benchmark, chord-regularized models with linear conditioning reuse data from five optimization campaigns and simulator-labeled rescalings to populate the Pareto front more densely. After low-fidelity screening, $93\,\%$ of $1.2$k generated candidates pass the benchmark's high-fidelity feasibility test. We release code and over $600$k designs for stellarator generation, surrogate training, and front-filling evaluation. These results support feasibility-aware latent learning for exploring constrained design trade-offs.
\end{abstract}
]
\thispagestyle{firststyle}

\section{Introduction}

Stellarators use three-dimensional magnetic fields to confine a hot
plasma in a toroidal volume. Their magnetic geometry can provide
confinement without requiring a large plasma current, but it also
creates a demanding design problem: shape changes affect several
physics properties at once~\cite{imbert2024introduction}.
In the boundary-optimization stage studied here, a truncated Fourier
series describes the plasma surface. Evaluating a candidate requires
solving an ideal magnetohydrodynamic (MHD) equilibrium and computing
confinement and stability metrics~\cite{freidberg1986ideal}.

Each optimization campaign is expensive and returns only part of the
available design trade-offs. Designers also need intermediate choices:
for example, different compromises between plasma compactness and
magnetic geometry favorable to coil design. Interpolation could reuse
existing campaigns to supply these choices, even when improving the
best objective value is not the goal. The difficulty is that the
feasible set is non-convex. In our dataset, families with near-planar
and strongly non-planar magnetic axes occupy separated regions of
the sampled design space despite overlapping physics metrics.

Learned latent generative models offer a route. A
variational autoencoder (VAE)~\cite{kingma2014autoencoding} or
conditional VAE (cVAE)~\cite{sohn2015learning} trained on feasible designs maps this
design space into a low-dimensional representation with a continuous
decoder. Interpolating between encoded designs then gives a sequence
of candidate boundaries, and optimization can search the same learned
space.

This depends on a property the cVAE objective never asks for: a
feasible \emph{interior}. Two endpoints $x_a, x_b$ can both be
strictly feasible while the decoded chord between them,
$D_\theta((1-\alpha) E_\phi(x_a) + \alpha E_\phi(x_b), y(\alpha))$ for
$\alpha \in [0, 1]$, passes through infeasible designs. Conditioning
compounds the problem. Interpolating the conditioning vector can request metric values
that the decoder cannot realize near the chord, further reducing
feasibility. Both failures
strike in the interior, which is exactly where interpolation and
latent search would be useful.

We introduce \method{}, a feasibility-regularized cVAE that addresses
these two difficulties. During training we sample pairs of feasible
designs, decode their latent interpolants, and penalize constraint
violations measured through a frozen physics surrogate. The path
between feasible endpoints therefore enters the training objective,
not only the endpoints themselves. At inference we test a self-consistent fixed-point map that
updates the conditioning vector using the decoded design's predicted
metrics, with optional constraint margins. This step improves the
fixed-aspect-ratio conditional-model comparison; the front-filling
experiment instead uses linear conditioning.

The two mechanisms together increase feasibility along sampled
decoded chords under simulator evaluation, and the same map drives a sampling-based
optimization procedure. With the same surrogate and evaluation budget, CMA-ES
returns feasible candidates from the \method{} latent and none
from the $81$-D design parameterization. This comparison demonstrates useful search in the learned
representation at matched surrogate access. It does not isolate
the effect of chord regularization from that of learning a latent
space, or establish that every native-space search method fails. Black-box optimizers such as the covariance matrix adaptation
evolution strategy (CMA-ES)~\cite{hansen2003reducing} remain
competitive on objective value once they run inside that latent.

The practical test is whether these models supply useful new design
choices. On a multi-objective stellarator benchmark, \method{} fills
gaps between five retained optimization campaigns, reducing the
largest aspect-ratio gap along the non-dominated front from $1.25$
to $0.51$. The benefit is better coverage of feasible trade-offs,
with a modest improvement in hypervolume.

Our contributions are:

\begin{itemize}[leftmargin=*]
    \item We introduce a training penalty on predicted constraint
    violations along decoded latent chords, and evaluate feasibility
    at sampled interior points with the simulator.
    \item We evaluate chord regularization in controlled examples
    and stellarator design, and ablate its interaction with
    self-consistent conditioning in fixed-aspect-ratio interpolation.
    \item We demonstrate front filling on the ConStellaration
    benchmark: learned interpolation adds feasible, non-dominated
    designs between retained campaigns without running another
    optimization campaign.
    \item We assemble reusable datasets: $100$k feasible boundaries
    spanning magnetic-axis geometry families, and the campaign and
    simulator-validated generated designs used to assess front filling.
\end{itemize}

\section{Related work}
\label{sec:related}

\paragraph{Latent regularization for constrained design.}
Deep latent generative models have been applied to inverse
engineering design across domains. The closest precedent for our
training-time contribution is COIL, which trains a VAE on feasible
solutions to bias constrained optimization toward valid
candidates~\citep{bentley2022coil}. We go further and penalize decoded
constraint violations along latent chords \emph{between} feasible
endpoints. Truss-VAE pairs a graph-based VAE with a property
predictor for metamaterial interpolation and inverse
design~\citep{zheng2023trussvae}. FlatVI regularizes latent
geometry so that straight paths approximate geodesics on the decoded
statistical manifold~\citep{palma2025flatvi}. Its criterion is
intrinsic to the manifold, where ours comes from an external
simulator. Decoder-uncertainty Bayesian
optimization~\citep{notin2021decoderuncertainty} and consistency-aware
latent search~\citep{boyar2024lsbo} make latent optimization robust to
an unreliable decoder, but neither targets pathwise constraint
satisfaction. Our self-consistent map shares the motivation of the
consistency-based methods. It differs in being a deterministic
fixed-point alignment of predicted metrics with the conditioning
vector. Latent-consistency methods instead address the discrepancy
between a latent code and its decoded-then-reencoded value. Tab.~\ref{tab:related-comparison}
(Appendix~\ref{sec:related-app}) summarizes the contrast.

\paragraph{Stellarator design with learned representations.}
Stellarator design has been driven by specialized optimization
pipelines built on ideal-MHD solvers
(e.g., \vmec~\citep{hirshman1983vmec,schilling2025numerics}, \desc~\citep{dudt2023desc}).
Large datasets such as QUASR~\citep{giuliani2024quasr,giuliani2025quasr}
and ConStellaration~\citep{cadena2025constellaration} provide plasma
boundaries for learning; ConStellaration also defines optimization
benchmarks. Near-axis studies have mapped families of
quasisymmetric~\citep{landreman2022mapping} and
quasi-isodynamic~\citep{rodriguez2026near} stellarators, while formal
measures quantify departures from quasisymmetry~\citep{rodriguez2022measures}.
Our quasi-isodynamic metric and constraints follow
ConStellaration~\citep{cadena2025constellaration}. Learned representations
have also been used for diffusion-based sampling of plasma
boundaries~\citep{padidar2025diffusion} and supervised dimensionality
reduction of boundary geometry~\citep{laia2026data}. These approaches are
complementary to ours. \method{} is not a new equilibrium solver or
direct optimizer; it asks whether the learned latent space carved
out from a feasibility-filtered dataset can be made navigable enough
that interpolation and sampling-based search are reliable on the
latent \emph{interior}, not only at training endpoints.

\section{Method}

\begin{figure*}[!t]
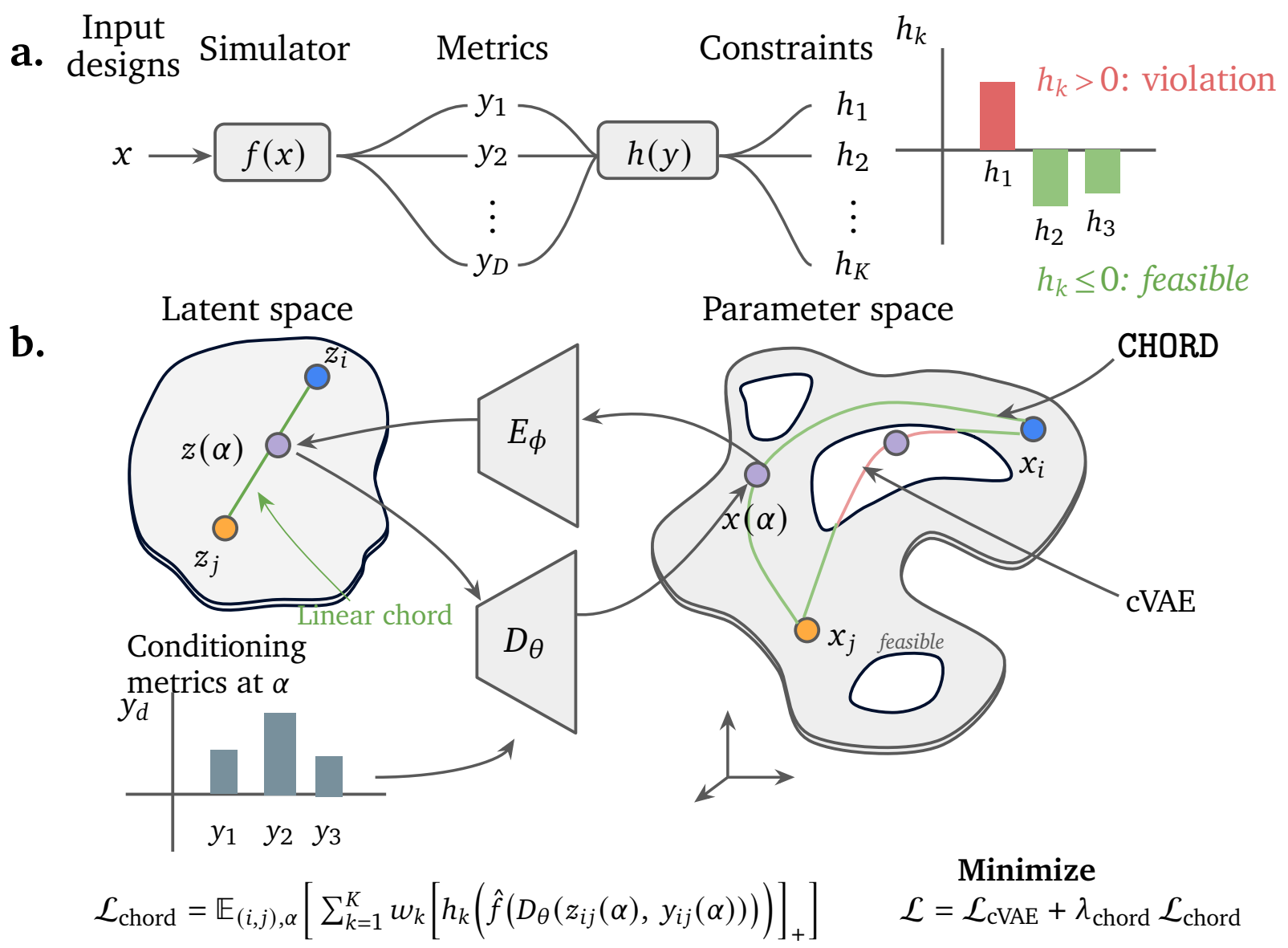

  \centering
  \includestandalone[width=0.80\linewidth]{figures/method}
  \caption{\textbf{Method.} \textbf{a}. Constraint functions on a
  simulator's output metrics define feasibility. \textbf{b}.
  \method{} penalizes predicted constraint violations at sampled
  points on decoded latent chords. Conditioning on feasible metric
  values alone does not constrain the decoded path.}
  \label{fig:method}
\end{figure*}
\label{sec:method}

\method{} penalizes predicted constraint violations along decoded
interpolants during training. At inference, it updates conditioning
to better match the decoded design's predicted metrics
(Fig.~\ref{fig:method}).

\paragraph{Setup.}
A design $x \in \X \subseteq \R^d$ is mapped by a simulator
$f : \X \to \R^M$ to a vector of physics metrics. Feasibility is
defined by $K$ constraint functions $h_k$ acting on those metrics,
with feasible set
$\F = \{ x \in \X : h_k(f(x)) \le 0\ \forall k \}$. A fixed,
deterministic subset of $f(x)$ is designated as the \emph{conditioning
vector} $y = y(x) \in \R^{d_y}$. It describes the properties requested
from the decoder, such as rotational transform or mirror ratio in
stellarator design. We train a conditional encoder and decoder on feasible designs.
The encoder receives $(x,y(x))$; for brevity we write its posterior
mean as $E_\phi(x) \in \Z$. The decoder is
$D_\theta : \Z \times \R^{d_y} \to \X$.

To interpolate, we connect the endpoint codes by a straight line
and decode points along it.
For endpoints $x_a, x_b \in \F$ with codes $z_a = E_\phi(x_a)$ and
$z_b = E_\phi(x_b)$, the linear latent chord is
$z(\alpha) = (1-\alpha)\,z_a + \alpha\,z_b$, $\alpha \in [0, 1]$,
decoded as $\tilde x(\alpha) = D_\theta(z(\alpha),\, y(\alpha))$ with
the conditioning trajectory $y(\alpha)$ chosen at inference time.
Reconstruction encourages small endpoint error but does not directly
control the pathwise violation
$V(\alpha) = \sum_k \relu{h_k(f(\tilde x(\alpha)))}$.

\paragraph{Training: chord regularization.}
The cVAE objective
\begin{equation}
  \begin{split}
  \mathcal{L}_{\text{cVAE}}
  &= \E_{q_\phi(z\mid x,y)}[-\log p_\theta(x \mid z,\, y)]\\
  &\quad + \beta\,\mathrm{KL}\bigl(q_\phi(z\mid x,y)\,\Vert\, p(z)\bigr)
  \end{split}
\end{equation}
balances reconstruction with latent regularity. Here KL is the
Kullback--Leibler divergence, which penalizes mismatch between the
encoder distribution $q_\phi(z\mid x,y)$ and the latent prior $p(z)$.
Neither term directly constrains decoded interpolants. We therefore
sample points between encoded feasible training designs and penalize
predicted constraint violations:
\begin{equation}
  \mathcal{L}_{\text{chord}}
  = \E_{(a,b),\,\alpha}\!\left[\sum_{k=1}^{K} w_k\,
    \relu{h_k\!\bigl(\hat f(\tilde x(\alpha))\bigr)}\right],
  \label{eq:chord-loss}
\end{equation}
evaluated through a frozen differentiable surrogate $\hat f$. The
surrogate supplies gradients without a simulator solve at each
training step.
For the fixed-aspect-ratio stellarator experiments, $w_k=1/K$,
so the penalty is the mean violation in physical metric units.
During
training we set $y(\alpha)$ to the linear blend of the two endpoints,
$y(\alpha) = (1-\alpha)\,y_a + \alpha\,y_b$. The self-consistent map of
Eq.~\eqref{eq:self-consistent} is used only at inference, never during
training. The full training objective is
\begin{equation}
  \mathcal{L} = \mathcal{L}_{\text{cVAE}} + \lambda\,\mathcal{L}_{\text{chord}},
  \label{eq:full-loss}
\end{equation}
with $\lambda$ ramped from $0$ after a reconstruction-and-KL warmup.
Gradients of $\mathcal{L}_{\text{chord}}$ flow through the decoder
and the surrogate but not the encoder; the simulator $f$ is reserved
for validation.

\paragraph{Inference: self-consistent conditioning with interior clamping.}
At test time the conditioning trajectory $y(\alpha)$ controls where
the decoder lands. A naive linear blend
$y(\alpha) = (1-\alpha)\,y_a + \alpha\,y_b$ can request
metric values that the decoder does not reproduce near $z(\alpha)$
(\S\ref{sec:stellarator-interp}). We therefore hold each latent point
fixed and update its conditioning: decode a design, predict its
metrics, and feed those predictions back as the next request.
This gives the fixed-point map
\begin{equation}
  y_{k+1}(\alpha) = \mathrm{clamp}\bigl(\,\hat f_y\!\bigl(D_\theta(z(\alpha),\, y_k(\alpha))\bigr)\bigr),
  \label{eq:self-consistent}
\end{equation}
where $\hat f_y$ projects the surrogate's full metric output onto the
$d_y$ conditioning components and $\mathrm{clamp}(\cdot)$ optionally
hard-projects the constrained components onto their
feasibility intervals shrunk by a small interior margin. The same
iteration is used in two regimes. For \emph{interpolation} between
encoded endpoints, $y_0(\alpha)$ is initialized to the linear blend
of $y_a$ and $y_b$ (\S\ref{sec:stellarator-interp}). For
\emph{sampling-based optimization}, $z \sim \mathcal{N}(0, I)$ is drawn
from the latent prior and $y_0$ is drawn with a task-specific
distribution biasing the iteration toward the desired metric tail
(\S\ref{sec:stellarator-opt}). We run a fixed number of iterations,
without assuming convergence. Even at a fixed point, clamped
components can differ from the decoder's predicted metrics. The
interior margin provides a buffer against surrogate error; we assess
feasibility of the final candidates with the simulator $f$.

\section{Results}
\label{sec:results}

We first isolate the training penalty on geometric examples, then
test stellarator interpolation, objective-driven search, and finally
the filling of gaps between optimization campaigns.

\subsection{Constructed examples}
\label{sec:synthetic}

The constructed problems ask whether penalizing decoded interpolants
helps when the feasible geometry is known exactly. C-shape makes the
interpolation failure visible in two dimensions; Shells tests a
compressed representation of four disconnected regions in 16 dimensions.
The forward model $y(x)$ is analytically available in both, so we
test the training penalty without surrogate error. We use the linear
blend $y(\alpha) = (1-\alpha) y_a + \alpha y_b$ and do not apply
self-consistent conditioning. Even here, blended endpoint metrics
need not correspond to a feasible design.

\begin{figure}[!t]
  \centering
  \includegraphics[width=0.50\linewidth]{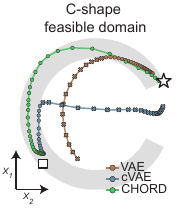}
  \caption{\textbf{2-D C-shape, decoded latent chord.} Between two
  feasible endpoints (star and square), cVAE cuts through the
  forbidden inner disc; \method{} arcs around it and stays feasible.}
  \label{fig:concept}
\end{figure}

\paragraph{C-shape: a 2-D mechanism illustration.}
The feasible set is a curved band shaped like a C: an annulus with
one angular sector removed. A feasible route between its arms must
follow the band around the central hole; a straight shortcut crosses
the forbidden disc. \method{} improves feasibility by bending decoded paths around this
hole: across $13$ seeds at $z=1$, mid-chord feasibility reaches
$\mathbf{0.65 \pm 0.03}$, compared with $0.21 \pm 0.02$ for cVAE.
It also reaches a held-out objective on the far side (Fig.~\ref{fig:concept}; full numbers in
App.~\ref{sec:cshape-app}).

\paragraph{Shells: four separated mechanism regions.}
In Shells, four groups of design variables represent alternative
mechanisms for producing a two-dimensional output. Feasibility
requires one mechanism to dominate, separating the design space
into four regions. The resulting output must also lie outside an inner
forbidden disc and within an outer bound, giving the shell-like
geometry. This tests whether regularization remains useful under
compression when no fully feasible path connects the regions.
At $z=1$,
\method{} improves sampled interior feasibility by $58$ percentage
points over cVAE and returns feasible solutions to a held-out
optimization task from $28/30$ starts, versus $20/30$.
Unconditional regularization can instead produce nearly constant
decoded paths: high sampled feasibility alone does not establish
useful interpolation. Appendix~\ref{sec:shells-app} gives the model, results,
and path diagnostics.

\subsection{Stellarator setup}
\label{sec:stellarator-setup}

We now turn to simulator-constrained stellarator design. We first
define the physical problem, then construct a fixed-aspect-ratio
dataset spanning two magnetic-axis geometry families to test
interpolation between different feasible shapes.

\begin{figure*}[t]
  \centering
  \includegraphics[width=1.0\linewidth]{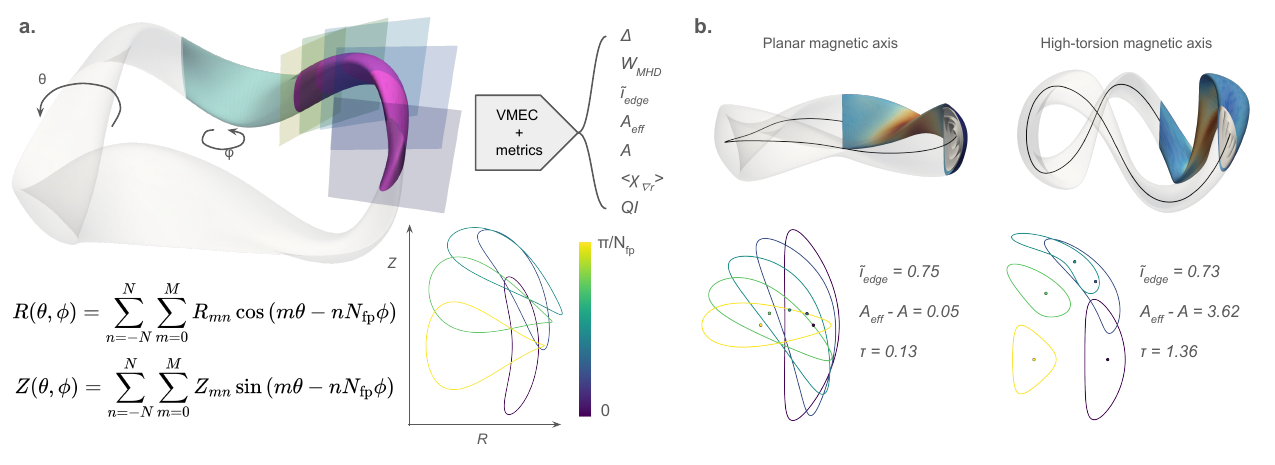}
  \caption{\textbf{Stellarator design pipeline.} Insets show a
  representative 3-D boundary, the resulting flux surfaces, and the
  magnetic axis. In \textbf{b}, the two designs have similar edge
  rotational transforms ($\iota_{\mathrm{edge}}=0.75$ and $0.73$)
  despite markedly different boundary shapes and magnetic-axis geometries.
  Axis non-planarity distinguishes the two sampled geometry families
  in Fig.~\ref{fig:stellarator-pca}.}
  \label{fig:stellarator-pipeline}
\end{figure*}

\paragraph{From boundary shape to physics.}
We design the outer plasma surface and assess the magnetic
equilibrium it supports. The equilibrium model assumes nested magnetic flux surfaces;
the magnetic axis is the central field line around which they lie.
A fixed-boundary calculation takes the outer surface, specified
profiles, and magnetic flux as inputs and solves for the equilibrium.
We optimize the boundary's Fourier coefficients and evaluate each
candidate with \vmec~\cite{hirshman1983vmec}
(Fig.~\ref{fig:stellarator-pipeline}). This stage evaluates magnetic
configurations; constructing and assessing coils is a further design
step~\cite{imbert2024introduction}.

\paragraph{Metrics and feasibility.}
The \textbf{rotational transform} $\iota$ measures how often a field
line winds around the torus's cross-section during one circuit
around the torus. The \textbf{mirror ratio} $\Delta$ measures the
variation in field strength on a flux surface, which affects particle
trapping and transport~\cite{banon2026optimizing}. The
\textbf{vacuum well} is an MHD-stability indicator. Our objective,
$\log_{10}\mathrm{QI}_{\mathrm{edge}}$, measures departure from
\textbf{quasi-isodynamic} magnetic geometry, which reduces the radial
drift of trapped particles; lower values are better. The subscript
``edge'' denotes evaluation on the outermost flux surface.
For the fixed-aspect-ratio experiments, feasibility means
$\iota_{\mathrm{edge}}\!\in\![0.725,0.775]$,
$\Delta_{\mathrm{edge}}\!\le\!0.4$, and vacuum well $\ge0$.
These are the study's constraints, rather than a complete assessment
of reactor performance. The front-filling task uses the separate
ConStellaration benchmark constraints
(Appendix~\ref{sec:front-filling-app}).

\paragraph{Geometry families.}
Without plasma current, rotational transform can arise from both
magnetic-axis torsion and flux-surface shaping. To test whether latent
interpolation can bridge separated regions of feasible designs, we
deliberately concentrate the dataset in near-planar-axis and strongly
non-planar-axis families, leaving a gap in axis-geometry coverage.
We distinguish these families using the effective-aspect-ratio
difference $A_{\mathrm{eff}}-A$, where $A$ is the major radius
divided by the minor radius. This difference is a proxy for axis non-planarity,
not a direct measurement of local torsion.
The families occupy separated regions of the sampled boundary
space while sharing much of the metric space.

\paragraph{Data generation.}
We collect designs from $204$ gradient-free optimization campaigns
at aspect ratio $AR=7$, minimizing $\log_{10}\mathrm{QI}_{\mathrm{edge}}$
under the constraints above (the \emph{broad-scan-loose} setting).
We bias the starting shapes to represent both axis families and
follow the augmented-Lagrangian approach of
ConStellaration~\citep{cadena2025constellaration}.
The optimization histories contain $5.09$\,M evaluated boundaries,
of which $1.01$\,M satisfy the constraints. All have three field
periods ($\mathrm{NFP}=3$): the boundary repeats three times around
the torus.

\begin{figure*}[t]
  \centering
  \includegraphics[width=\linewidth]{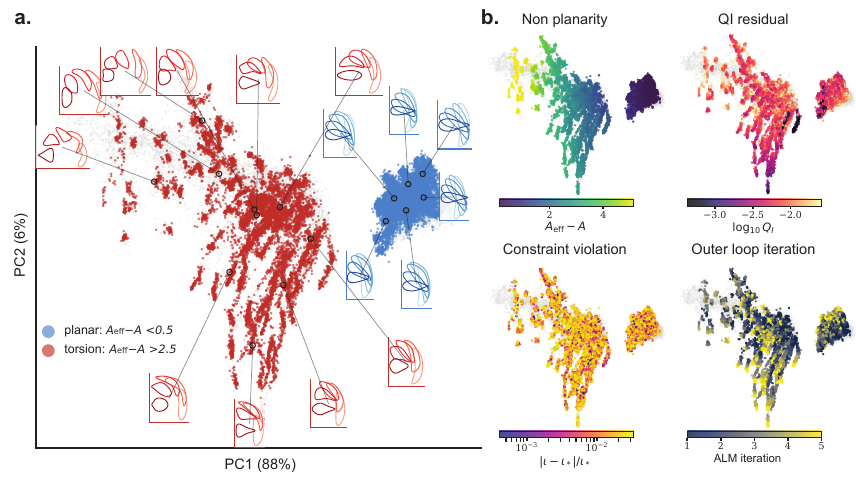}
  \caption{\textbf{The two stellarator basins, projected on the data
  PCA.} PC$_1$ is dominated by $A_{\mathrm{eff}}\!-\!AR$ and
  separates near-planar-axis from strongly non-planar-axis
  designs; both families span the $\log_{10}\mathrm{QI}_{\mathrm{edge}}$
  tail.}
  \label{fig:stellarator-pca}
\end{figure*}

Generating this corpus cost about $73$k vCPU-hours. The learned
models reuse it across interpolation and search queries; we draw a
$100$k row-stratified subset of the feasible corpus for training,
validation, and test, split by row identity. These boundaries use
\vmec's \texttt{very\_low\_fidelity} preset; the later front-filling
experiment uses the benchmark's high-fidelity evaluation.
Appendices~\ref{sec:stellarator-setup-app} and
\ref{sec:stellarator-opt-app} give the generation protocol and
separate one-time from per-query costs.
Figure~\ref{fig:stellarator-pca} shows how the two geometry families
separate in a projection onto the dataset's first two principal components.

\paragraph{A reusable dataset spanning axis geometries.}
The released $100$k-boundary corpus contains both near-planar and
strongly non-planar axis designs, with boundary coefficients,
physics metrics, geometry-family labels, and saved training,
validation, and test assignments for reproducible comparisons.
The samples come from optimization histories and do not represent
a uniform survey of feasible stellarators. Appendix~\ref{sec:stellarator-setup-app}
gives the corpus composition; access is described under
\hyperref[sec:data-availability]{Data and code availability}.

\paragraph{Training the four model variants.}
To separate the effects of conditioning and chord regularization,
we train four model families: an unconditional \textbf{VAE}, a
conditional \textbf{cVAE}, the chord-regularized cVAE
\textbf{\method{}}, and an unconditional chord-regularized VAE
\textbf{\method{}-unc}. All four share a residual multilayer perceptron
and training schedule; we vary the latent dimension and repeat
training with three seeds. Each model takes about $0.5$ GPU-hours
to train. The conditional models receive six geometry and physics metrics,
including the QI objective and the three constrained metrics.
The frozen surrogate is trained on $90\,\%$ of the full $5.09$\,M-row
archive, including infeasible designs, with the remainder held out
for validation. Appendix~\ref{sec:stellarator-setup-app} gives the
conditioning vector, training settings, and reconstruction quality.

\subsection{Stellarator cross-basin interpolation}
\label{sec:stellarator-interp}

\paragraph{Cross-basin protocol.}
The first test asks whether decoded paths remain feasible between
the two geometry families, which we call the \textbf{planar-axis}
and \textbf{torsion-axis} basins. Here ``basin'' labels a region of
the dataset; it does not establish topological disconnection of the
full feasible set. A cVAE reconstructs both cleanly, but the straight latent
chord between them usually passes through infeasible interiors.

We use the same $30$ planar$\leftrightarrow$torsion endpoint pairs
from the validation split across model variants, latent dimensions,
and training seeds. We
encode the endpoints, decode a $21$-step linear chord
$z(\alpha) = (1-\alpha)\,z_a + \alpha\,z_b$ for each pair, and
validate \emph{every} decoded boundary on real \vmec. The sweep covers the four model variants, four latent dimensions,
three training seeds, and two conditioning schemes. We report
mid-segment feasibility over eight sampled points in the middle of
each path. Appendix~\ref{sec:stellarator-interp-app} gives the exact
grid, its sensitivity analysis, and endpoint feasibility separately.

\begin{figure*}[t]
  \centering
  \begin{minipage}[c]{0.64\textwidth}
    \includegraphics[width=\linewidth]{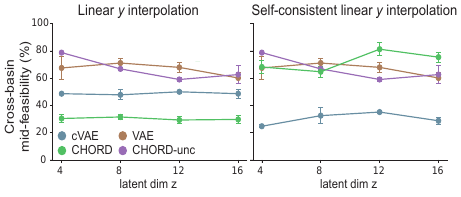}
  \end{minipage}\hfill
  \begin{minipage}[c]{0.33\textwidth}
    \caption{\textbf{Cross-basin mid-feasibility under \vmec
    validation.} Mean $\pm$ SE over three seeds, $30$ pairs $\times$
    $8$ mid-$\alpha$ points each. Under linear $y$ (left) \method{}
    trails the cVAE; under self-consistent $y$ (right) it reaches
    $65\text{--}81\,\%$ while the cVAE falls to
    $25\text{--}35\,\%$.}
    \label{fig:interpolation-nums}
  \end{minipage}
\end{figure*}

\paragraph{Two conditioning schemes.}
We compare inference with and without self-consistency to test how
it interacts with the training penalty.
For conditional decoders, we either linearly blend the endpoint
metrics or start from that blend and apply five self-consistent
updates (Eq.~\eqref{eq:self-consistent}). Unconditional decoders
receive no conditioning, so the two schemes coincide for them.

\begin{figure*}[t]
  \centering
  \includegraphics[width=1.0\linewidth]{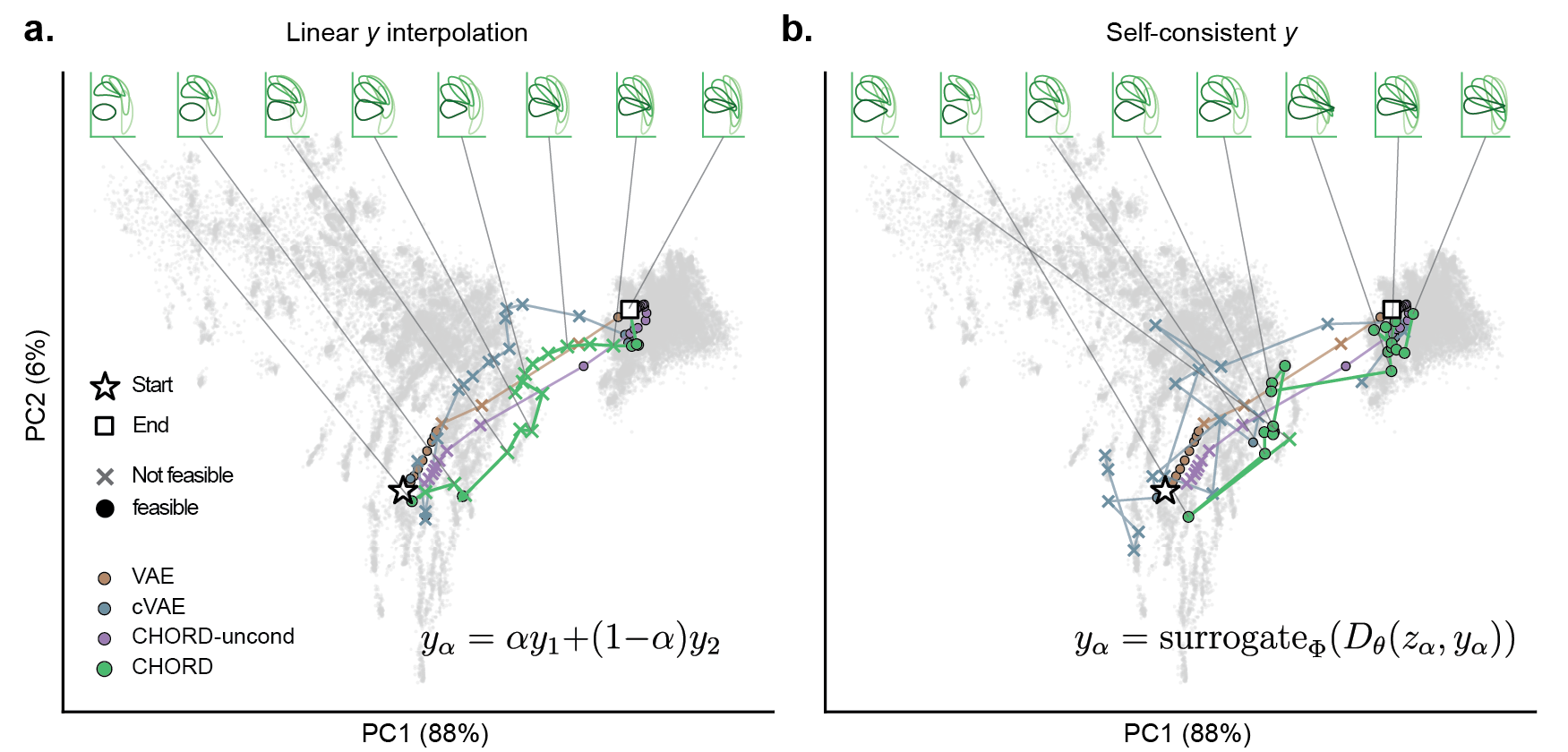}
  \caption{\textbf{Decoded latent chords for one cross-basin pair,
  projected to the data PCA.} \textbf{a.}~Under linear $y$, both
  conditional models have infeasible sampled points (crosses).
  \textbf{b.}~With self-consistent $y$, all sampled points on
  \method{}'s chord are feasible in this example. Insets show the boundary morphing from a planar to a
  torsion-axis configuration.}
  \label{fig:stellarator-bars}
\end{figure*}

\paragraph{Training and inference work together.}
For the conditional models, the improvement comes from combining
the training penalty with self-consistent inference. Together they
reach $65\text{--}81\,\%$ mid-segment feasibility across the tested
latent dimensions, compared with $48\text{--}50\,\%$ for the
unregularized cVAE with linear conditioning
(Fig.~\ref{fig:interpolation-nums}). Removing either ingredient
reduces feasibility below that baseline. These ablations establish
an interaction in this experiment without identifying a geometric
cause. Figure~\ref{fig:stellarator-bars}b shows a representative path.

Although the unconditional VAE and \method{}-unc remain competitive in
mid-segment feasibility ($59\text{--}79\,\%$, leading at the two smallest
latent dimensions), their intermediate designs cluster closer to the
decoded endpoints than \method{}'s (illustrated for one pair in
Fig.~\ref{fig:stellarator-bars}). Across the tested dimensions and
seeds, the mean distance from the nearer decoded endpoint is
$19\,\%$ and $20\,\%$ of the decoded endpoint separation for VAE and
\method{}-unc, respectively, versus $34\,\%$ for \method{} with
self-consistency. Thus \method{} combines high feasibility with less
endpoint concentration, a distinction that feasibility alone misses.
Appendix~\ref{sec:stellarator-interp-app} defines this post-hoc
boundary-coefficient distance diagnostic and reports the full comparison.

\paragraph{Feasibility under prescribed metric targets.}
With a requested metric profile held fixed during the conditioning
updates, \method{} retains $77.9$ to $81.7\,\%$ mid-segment feasibility
across three profiles, compared with $14.6$ to $40.8\,\%$ for the
unregularized cVAE in a comparison using one training seed. When asked to retain
the better endpoint's QI, \method{} more than halves mean QI
deterioration relative to the unsteered VAE: $0.10$ versus $0.22$ dex
in a post-hoc one-sided analysis. This result shows that \method{}
can limit the loss of a favorable endpoint property while exploring
intermediate designs. Appendix~\ref{sec:controllability-app} reports
the full feasibility and target-tracking comparison.

\paragraph{Alternative paths and generative models.}
Spherical interpolation (slerp), which follows a curved latent path,
improves the unconditional VAE but adds no gain for \method{}.
On the same seed-0 checkpoint and $30$ endpoint pairs, \method{}
with linear interpolation and self-consistent conditioning reaches
$81.7\,\%$ feasibility, compared with $71.7\,\%$ for VAE+slerp.
This matched comparison does not establish a multi-seed advantage.
A conditional flow-matching baseline~\cite{lipman2023flowmatching}
reaches $59.9\pm4.4\,\%$ with linear interpolation across three
seeds. Adding the chord penalty and self-consistent inference does
not improve on this baseline. Appendix~\ref{sec:baselines-app}
reports the full path and flow-matching ablations.

\subsection{Stellarator latent optimization}
\label{sec:stellarator-opt}

Interpolation follows paths between chosen endpoints. We next ask
whether the learned space also supports searching for better
objective values and sampling a broader set of feasible alternatives.

\begin{figure*}[t]
  \centering
  \begin{minipage}[c]{0.71\linewidth}
    \includegraphics[width=\linewidth]{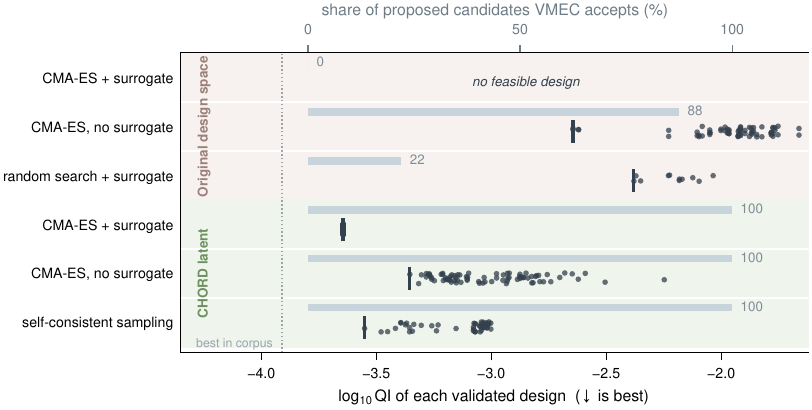}
  \end{minipage}\hfill
  \begin{minipage}[c]{0.26\linewidth}
    \caption{\textbf{Search outcomes in native and learned coordinates.}
    Each dot is a \vmec-validated design. Bars (top axis) give the
    share of proposed candidates \vmec accepts: $0$ to $88\,\%$ in the
    original parameterization, $100\,\%$ for the methods shown in the
    \method{} latent. Depth then depends on the procedure.}
    \label{fig:search-space}
  \end{minipage}
\end{figure*}

\paragraph{Search and sampling protocols.}
All methods minimize $\log_{10}\mathrm{QI}$ under the broad-scan-loose
constraints at $z=16$. The surrogate-assisted methods share one frozen
surrogate, with $300$k surrogate evaluations for search and a
budget of $50$ designs for \vmec validation. Self-consistent sampling
uses a further $10$k surrogate evaluations to rank its final decoded
designs. The surrogate-free CMA-ES baselines use $200$ \vmec calls per
trial; trial counts are given in Appendix~\ref{sec:stellarator-opt-app}.

As an alternative to optimizing latent coordinates, we sample codes
and adapt their conditioning toward favorable metrics.
\textbf{Self-consistent sampling} draws $10{,}000$ codes from the
latent prior, initializes conditioning toward low QI, and applies
$30$ self-consistent updates with interior clamping. We rank the
decoded designs by surrogate QI and validate the best $50$ with
\vmec. The inference loop costs about $30$\,s of CPU; initialization
and clamp settings are given in Appendix~\ref{sec:stellarator-opt-app}.

\paragraph{The learned representation improves surrogate-assisted CMA-ES.}
Changing the search coordinates makes the same optimizer more
successful at finding feasible designs. Surrogate-assisted CMA-ES
returns $100\,\%$ feasible candidates in the \method{} latent,
compared with $0\,\%$ in the original $81$-D parameterization,
with $50$ designs validated in each case
(Fig.~\ref{fig:search-space}). These runs share the surrogate and
evaluation budget but use parameterization-specific initializations
and step sizes (Appendix~\ref{sec:stellarator-opt-app}). Other
native-space methods reach up to $88\,\%$ feasibility. All latent
search methods shown in Fig.~\ref{fig:search-space} reach
$100\,\%$; the cycle-penalty test below is a separate comparison.

The native-space failure is consistent with surrogate extrapolation:
those candidates lie far from the training data, where surrogate
scores are unusually favorable, while latent candidates stay
near the training distribution. \vmec fails to converge on the
native-space candidates; this does not establish that no equilibrium
exists. Appendix~\ref{sec:stellarator-opt-app} gives the distance analysis.

\paragraph{Self-consistent sampling improves objective value and coverage.}
Adapting the conditioning improves the objective value of sampled
designs: on the same latent draws, the best $\log_{10}\mathrm{QI}$
improves from $-2.04$ with fixed conditioning to $-3.55$ with
self-consistency, while retaining $100\,\%$ feasibility. All $150$
candidates validated across three training seeds are feasible.
Latent CMA-ES reaches a better best value of $-3.66$, but its
candidates cluster tightly. Sampling provides a wider range of both
objective values and designs: its range of $\log_{10}\mathrm{QI}$
is $0.55$, compared with $0.001$ for CMA-ES
(Appendix~\ref{sec:stellarator-opt-app}). Thus the same representation
supports focused optimization and exploration of feasible alternatives.

\begin{figure*}[t]
  \centering
  \includegraphics[width=0.70\linewidth]{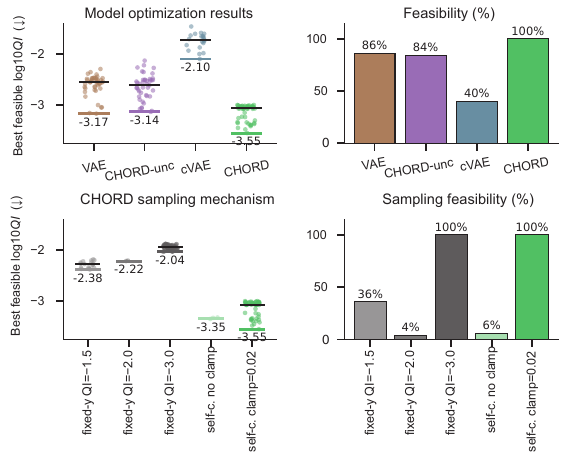}
  \caption{\textbf{Self-consistent latent sampling: model and mechanism
  ablations} ($z=16$, top-$50$ \vmec validations per cell).
  \textbf{Top:} best feasible $\log_{10}\mathrm{QI}$ and feasibility
  rate across architectures. \textbf{Bottom:} fixed-$y$ conditioning
  at three QI targets against self-consistent inference, with and
  without interior clamping.}
  \label{fig:optimization-nums}
\end{figure*}

\paragraph{An interior margin improves sampling feasibility.}
To identify what makes the sampler effective, we remove interior
clamping and chord regularization in turn. Without the clamp in Eq.~\eqref{eq:self-consistent}, only
$6\,\%$ of the selected candidates pass \vmec
(Fig.~\ref{fig:optimization-nums}, lower row).
The matched margin sweep identifies slightly negative vacuum well
as the failure in all $47$ rejected candidates; neither rotational
transform nor mirror ratio is the binding constraint.
With margin $0.02$, all $50$ selected candidates pass.
The clamp acts on the conditioning values, so its bounds do not
guarantee the same bounds on the decoded design's simulated metrics. This result supports using an interior margin for the tested
self-consistent sampler.

\paragraph{Chord regularization improves the conditional sampler.}
Running the same self-consistent + clamping pipeline on an unregularized
cVAE produces only $20/50$ \vmec-feasible candidates and a best of
$-2.10$ (Fig.~\ref{fig:optimization-nums}, upper row). Thus the inference procedure alone does not recover the performance
of the regularized model. Unconditional sampling also gives competitive objective values,
with about $85\,\%$ feasibility. Among these architecture ablations,
the full \method{} pipeline combines the best objective value with
$100\,\%$ feasibility; the numerical breakdown is given in
Appendix~\ref{sec:stellarator-opt-app}.

\paragraph{Cycle consistency does not improve this search.}
Motivated by methods that use decoder uncertainty to guide latent
search, such as that of Notin et al.~\cite{notin2021decoderuncertainty},
we test whether an additional consistency penalty improves optimization
on the frozen \method{} representation. Our deterministic penalty
measures how far a decoded design re-encodes from its starting latent
code. At the tested weight, it yields nine distinct candidates, five
of them feasible; unpenalized latent CMA-ES and self-consistent sampling
each yield $50$ feasible designs and reach better QI. Thus the
additional cycle penalty offers no benefit in this comparison.
Appendix~\ref{sec:baselines-app} explains the distinction from Bayesian
decoder uncertainty and gives the protocol and results.

\subsection{Filling a public benchmark front}
\label{sec:front-filling}

The preceding tests hold aspect ratio fixed. We now let it vary and
ask whether interpolation can add feasible trade-off choices between
expensive optimization campaigns.

\begin{figure*}[t]
  \centering
  \includegraphics[width=\linewidth]{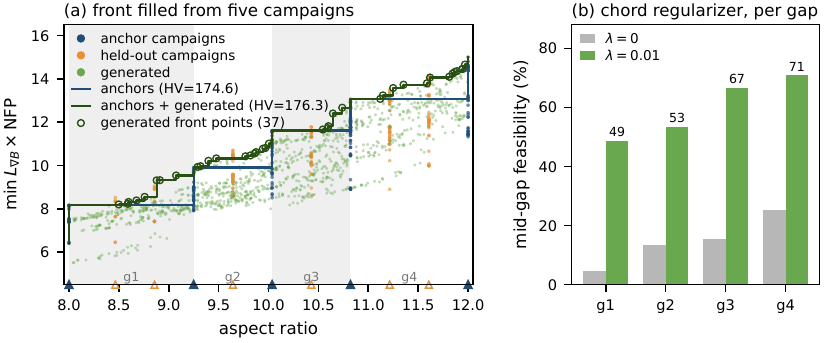}
  \caption{\textbf{Generated designs fill gaps between five retained
  campaigns.} \textbf{(a)} The combined front contains $37$ generated
  designs (outlined) and $14$ surviving anchor designs. The area
  between the staircases is the added hypervolume. Six held-out
  campaigns (orange) provide comparison data.
  \textbf{(b)} Low-fidelity mid-gap feasibility with the chord
  regularizer off and on.}
  \label{fig:front-filling}
\end{figure*}

\paragraph{Setup.}
The ConStellaration MHD-stable benchmark trades off two objectives:
lower aspect ratio $A$ (major radius divided by minor radius), which
favors a compact plasma, and larger minimum magnetic-gradient scale
length $\min L_{\nabla B}\!\times\!\mathrm{NFP}$ (in units of the minor radius), a proxy for
coil-to-plasma separation~\cite{cadena2025constellaration}.
A design is non-dominated if no other feasible design improves one
objective without worsening the other. The resulting front offers
choices between the two objectives; hypervolume measures its
coverage relative to a fixed reference point.
We study eleven expensive, gradient-free optimization campaigns
within this benchmark. Our aim is to add feasible intermediate
choices using a subset of those campaigns' data. We keep five as anchors, spanning aspect ratio
$8$ to $12$, and hold out the six that lie between them. For each of
the four gaps we train a \method{} model using the two campaigns
that bracket it and simulator-labeled rescalings of their boundaries.
These augmented designs include aspect ratios inside the gap.
The split excludes designs originating from the held-out campaigns,
rather than excluding all training data at intermediate aspect ratios.

These models use linear conditioning: self-consistent inference does
not recover feasibility in the gap where it was tested
(Appendix~\ref{sec:front-filling-app}). We screen generated
designs at low fidelity, then evaluate $1{,}200$ of the $8{,}361$
low-fidelity-feasible candidates at high fidelity. The feasibility
rate below describes this selected subset, not all generated proposals.

\paragraph{Generated designs fill the gaps at high fidelity.}
The selected designs populate the gaps with a high feasible yield:
$93\,\%$ of the $1{,}200$ evaluated candidates pass the benchmark's
high-fidelity test. This permits $1\,\%$ normalized constraint
violation; $81.4\,\%$ also satisfy the strict nominal thresholds. In a separate geometric-rescaling control at $\Delta A=-0.5$,
none of nine rescaled neighboring solutions is feasible.
The main benefit is a denser set of trade-off choices: the largest
adjacent aspect-ratio gap along the front decreases from $1.25$ to
$0.51$, with generated designs contributing in all four gaps
(Fig.~\ref{fig:front-filling}a). Hypervolume increases modestly,
from $174.6$ to $176.3$. Appendix~\ref{sec:front-filling-app} gives
the non-dominated counts, comparisons with the held-out campaigns
and broader benchmark, and sensitivity to feasibility tolerance
and numerical resolution.

\paragraph{The chord regularizer raises mid-gap feasibility in all four gaps.}
To isolate the training penalty's contribution, we compare models
with and without it under the same linear conditioning.
Low-fidelity feasibility near the chord midpoint improves in every
gap; in the hardest gap it rises from $4.6$ to $48.7\,\%$
(Fig.~\ref{fig:front-filling}b). Difficulty does not increase
monotonically with gap width.
The hardest gap also has the lowest fraction of feasible designs
in the source corpus. App.~\ref{sec:front-filling-app}
gives the gap geometry, per-gap results, and the negative results.

\paragraph{Data for evaluating front filling.}
The accompanying dataset records all $1{,}200$ high-fidelity
evaluations, including designs that fail the benchmark constraints,
with boundary coefficients, gap labels, interpolation positions,
and simulated metrics. A separate table contains $3{,}572$
campaign-history rows and their roles in the comparison. Together
these allow the filled front to be rescored and feasibility to be
assessed under different thresholds. They complement the
fixed-aspect-ratio corpus: one tests interpolation across axis
geometries, the other tests the recovery of useful trade-off choices
between optimization campaigns.

\section{Discussion}
\label{sec:discussion}

\method{} improves feasibility at sampled latent interior points
and supports constrained search in the experiments reported here.
The roles of its two components depend on the task.
For fixed-aspect-ratio stellarator interpolation, the chord
regularizer and self-consistent inference work together: removing
either reduces feasibility below the unregularized cVAE with linear
conditioning. The constructed examples and front-filling experiment
use linear conditioning instead; self-consistent inference fails in
the tested front-filling setting. Thus the evidence supports a
task-dependent inference choice.

The front-filling experiment illustrates the intended use: reuse
expensive campaign data to explore feasible intermediate designs.
The generated designs reduce the largest aspect-ratio gap along
the front from $1.25$ to $0.51$, providing intermediate choices
before further physics and coil assessments. The modest hypervolume
gain is consistent with that aim; finding
a new best objective value is not required for interpolation to be
useful. The approach still requires labeled augmentation, model
training, and simulator validation, so it complements the original
campaigns rather than eliminating their cost.

\paragraph{Candidate selection is robust to changing the ranking surrogate.}
The surrogate is frozen during decoder training. This prevents
joint adaptation of the surrogate and generator, but does not prevent
the generator from exploiting surrogate errors. On $509$k held-out
boundaries it predicts feasibility at $96.8\,\%$ accuracy with
$\log_{10}\mathrm{QI}$ RMSE $0.032$. Using three independently trained alternatives to rank the same
candidate pool leaves $199$ of $200$ validated
candidates feasible and best QI within $0.03$ dex. The selected designs do change: only $12$ to $17$ of each top-$50$
match the original picks. A four-member ensemble raises feasibility prediction to
$97.6\,\%$ (App.~\ref{sec:surrogate-app}).

\paragraph{Limitations.}
(i) Pathwise regularization penalizes sampled chord points but does
not give a hard feasibility guarantee. (ii) Training
requires a simulator-labeled corpus and a surrogate $\hat f$.
Their one-time costs can be substantial, so the computational benefit
depends on reusing them across design queries. (iii) Surrogate
feasibility can disagree with simulator feasibility; ensembling
improves held-out calibration
(App.~\ref{sec:surrogate-app}). (iv) The fixed-aspect-ratio corpus
uses a row-level split of correlated optimization trajectories.
Its validation results measure in-distribution performance, not
generalization to unseen campaigns. The generative and surrogate
splits are separate; we do not report a holdout shared by both models. (v) The front-filling models
use simulator-labeled augmentation inside the gaps, and their
high-fidelity feasibility rate is conditional on low-fidelity
screening.

\section*{Data and code availability}
\label{sec:data-availability}

Data and trained model checkpoints are available at
\url{https://huggingface.co/datasets/proxima-fusion/chord-paper},
revision \texttt{28f8453}. The release contains the $100$k-boundary
corpus spanning magnetic-axis geometries, with saved splits;
a $500$k-boundary dataset for surrogate training, sampled from the
historical $5.09$M-evaluation archive; the campaign comparison table;
and $1{,}200$ high-fidelity front-filling evaluations. It also includes
boundary objects for the $1{,}120$ benchmark-feasible generated
designs, model checkpoints with normalization data, and the prepared
per-gap generator training, validation, and test splits. The dataset
card documents metric units, feasibility thresholds, provenance,
file hashes, and checks of the released model loaders.

The full historical training datasets are available from the
corresponding author on request. The reference implementation,
loading examples, and reproduction instructions are available at
\url{https://github.com/proximafusion/chord}.

\begin{ack}
We thank Tobias Weißberg and Florian Bernard for early discussions on
latent space learning in this context.
This work was independently funded by Proxima Fusion, and supported by
the BMBF grant FUSKI (FKZ: 13F1012A).
\end{ack}

\begingroup
\small
\bibliographystyle{unsrtnat}
\bibliography{references}
\endgroup
\clearpage

\appendix

\section{Technical Appendix}

\subsection{Why feasible endpoints do not ensure feasible interpolation}

Accurate endpoint reconstruction leaves the route between endpoints
unconstrained. Consider a nonconvex feasible set $\F\subset\R^d$
and feasible points $x_a,x_b\in\F$ whose Euclidean segment intersects
$\X\setminus\F$. If the decoder maps the latent chord approximately
to that segment, some intermediate designs will be infeasible even
when the endpoints are reconstructed exactly. The VAE objective
controls reconstruction and prior regularity; it does not directly
control $h_k(f(D_\theta(z)))$ along the segment.

Dense sampling reduces the unchecked gaps along a path. If
$g_k(\alpha)=h_k(f(D_\theta(z_{ij}(\alpha),y_{ij}(\alpha))))$ is
$L$-Lipschitz along a segment, its variation is bounded by $L$ times
the change in $\alpha$. On a grid with spacing $\Delta$,
violations between grid points are therefore bounded by the maximum
grid violation plus $L\Delta$. This motivates dense interpolation
checks, but $L$ is generally unknown and may vary across designs,
so the sampled results do not provide a global guarantee.

\subsection{Algorithms}

Training updates the decoder using violations on decoded interpolants;
inference holds the model fixed and updates the conditioning.
Tables~\ref{tab:chord-training} and~\ref{tab:selfconsistent-sampling}
summarize these two procedures.

\begin{table*}[t]
  \caption{Training \method{}.}
  \label{tab:chord-training}
  \centering
  \small
  \begin{tabular}{p{0.95\linewidth}}
    \toprule
    1. Sample a minibatch of feasible designs $(x, y)$. \\
    2. Compute cVAE reconstruction and KL losses. \\
    3. Sample endpoint pairs $(a, b)$ and interpolation coefficients $\alpha$. \\
    4. Encode endpoints, form $z(\alpha)$, and decode $\tilde{x}(\alpha) = D_\theta(z(\alpha), y(\alpha))$. \\
    5. Evaluate surrogate metrics $\hat{f}(\tilde{x}(\alpha))$. \\
    6. Compute ReLU constraint violation and update using Eq.~\eqref{eq:full-loss}. \\
    \bottomrule
  \end{tabular}
\end{table*}

\begin{table*}[t]
  \caption{Inference: self-consistent latent sampling.}
  \label{tab:selfconsistent-sampling}
  \centering
  \small
  \begin{tabular}{p{0.95\linewidth}}
    \toprule
    1. Sample $N$ latent codes $z \sim \mathcal{N}(0, I)$ and initial conditioning vectors $y_0$ with task-specific bias toward the desired metric tail. \\
    2. Iterate the self-consistent fixed-point map of Eq.~\eqref{eq:self-consistent} for $K$ steps, with interior clamping on the constrained components of $y$. \\
    3. Decode every $(z, y_K)$ pair and surrogate-score the resulting designs. \\
    4. Take the top-$K_{\mathrm{VMEC}}$ surrogate-feasible candidates by predicted objective. \\
    5. Validate the top-$K_{\mathrm{VMEC}}$ designs with the true simulator and report feasibility-validated objective values. \\
    \bottomrule
  \end{tabular}
\end{table*}

\subsection{Comparison with related methods}
\label{sec:related-app}

Table~\ref{tab:related-comparison} compares the mechanisms and
demonstrated tasks of the methods discussed in \S\ref{sec:related}.
The checkmarks identify reported applications, not matched performance tests.

\begin{table*}[t]
  \caption{Comparison to related directions. \method{} is
  distinguished by an explicit penalty on simulator-defined constraint
  violation along decoded linear latent interpolants, paired with a
  self-consistent fixed-point map for the conditioning vector at
  inference time. Checkmarks denote tasks demonstrated in the cited studies.}
  \label{tab:related-comparison}
  \centering
  \small
  \setlength{\tabcolsep}{5pt}
  \begin{tabular}{@{}lcclc@{}}
    \toprule
    Work & Interp. & Latent opt. & Key mechanism & Stellarator design \\
    \midrule
    COIL~\citep{bentley2022coil} & -- & \checkmark & Feasible training data & -- \\
    Truss VAE~\citep{zheng2023trussvae} & \checkmark & \checkmark & Property prediction & -- \\
    FlatVI~\citep{palma2025flatvi} & \checkmark & -- & Latent geometry & -- \\
    Decoder uncertainty~\citep{notin2021decoderuncertainty} & -- & \checkmark & Epistemic uncertainty & -- \\
    LCA-LSBO~\citep{boyar2024lsbo} & -- & \checkmark & Latent consistency & -- \\
    \textbf{\method{}} & \checkmark & \checkmark & \shortstack[l]{Chord penalty +\\metric alignment} & \checkmark \\
    \bottomrule
  \end{tabular}
\end{table*}

\subsection{C-shape: setup and results across training seeds}
\label{sec:cshape-app}

\paragraph{Geometry.}
The 2-D feasible region is a thin annulus with a forbidden sector around
the $+x_1$ axis. Writing $r = \|x\|_2$ and $\theta = \mathrm{atan2}(x_2, x_1)$,
\begin{equation*}
\begin{split}
\mathcal{F}_C = \{ x \in \R^2 : &\ r_{\text{in}} \le r \le r_{\text{out}},\
  |\theta| \ge \theta_{\min} \},\\
&\ r_{\text{in}}=0.85,\ r_{\text{out}}=1.15,\\
&\ \theta_{\min}=0.45\,\mathrm{rad}.
\end{split}
\end{equation*}
The region is path-connected but non-convex. For the endpoint pair
used in Fig.~\ref{fig:concept}, direct interpolation crosses the
forbidden inner disc.

\paragraph{Objectives.}
We generate training data by minimizing $f_{\text{obj}}(x)=-x_1$,
which concentrates designs near the two $+x_1$ tips of the C.
At evaluation, we instead minimize $f_{\text{opt}}(x)=+x_1$, whose
minimum is $-1.15$ at $(-1.15,0)$, on the opposite side of the C.
This tests search away from the region favored by data generation.
The new objective is not a component of the conditioning vector.

\paragraph{Conditioning, architecture, training.}
The cVAE and \method{} use conditioning vector
$y(x)=(\|x\|,\ |\theta|,\ x_2^2)$. We compare three models with
three hidden layers of $64$ units: an unconditional VAE, an
unregularized cVAE, and \method{} with $\lambda_{\text{chord}}=1$
and a $30$-epoch warmup. Each model has one latent dimension and
trains for $200$ epochs with cyclical KL annealing. We repeat
training with $13$ random seeds.

\paragraph{Data generation.}
For each half of the C, we run $20$ CMA-ES searches from randomly
selected feasible points, using population $8$, $30$ generations,
and initial step size $\sigma_0=0.15$. The search minimizes
$f_{\text{obj}}(x)+50\cdot\mathrm{ReLU}(\max_k h_k(x))$.
Pooling the trajectories and retaining feasible points yields
approximately $6.4$k designs per seed, balanced across the two halves.

\paragraph{Diagnostics.}
Identifiers in parentheses match the archived analysis and figure labels.
\begin{description}[leftmargin=1em,style=nextline,itemsep=2pt,topsep=2pt]
  \item[Cross-half interpolation (T1).]
    We encode validation endpoints from opposite halves, linearly
    interpolate $(z,y)$ over $21$ values of $\alpha$, and decode the
    path. We report the feasible fraction in the middle interval
    $\alpha\in[0.3,0.7]$, averaged over $30$ random endpoint pairs.
  \item[Novel-objective optimization (T7).]
    We minimize $f_{\text{opt}}$ with Adam in $(z,y)$, starting from
    the $30$ training designs with lowest $f_{\text{obj}}$.
    Each search uses $400$ iterations, a feasibility penalty of
    weight $10$, and a soft penalty when a standardized conditioning
    component has magnitude above $4$. We report the best
    feasible objective and the number of successful starts.
  \item[Sampling-objective optimization (T3).]
    The same search minimizes $f_{\text{obj}}$ instead. This checks
    whether the learned representation supports optimization of the
    objective used to generate the training data.
\end{description}

\paragraph{Endpoint pair for Fig.~\ref{fig:concept}.}
We select $A$ as the upper-half validation point with maximum $x_1$
and $B$ as the lower-half validation point with minimum $x_1+x_2$.
The straight chord $A\!\to\!B$ crosses the forbidden inner disc.

\paragraph{Results across training seeds.}
The regularizer improves both interpolation feasibility and search
for the new objective (Table~\ref{tab:cshape-13seeds}).

\begin{table*}[t]
  \centering
  \small
  \caption{C-shape results at $z=1$, mean $\pm$ SE over $13$ training
  seeds. Interpolation is scored by mid-segment feasibility. Each
  optimization diagnostic reports the best feasible objective and
  the number of successful starts out of $30$. The novel objective
  has minimum $-1.15$ at $(-1.15,0)$.}
  \label{tab:cshape-13seeds}
  \begin{tabular}{lccccc}
    \toprule
    & \shortstack{Mid-segment\\feasibility $\uparrow$}
    & \multicolumn{2}{c}{Novel-objective search}
    & \multicolumn{2}{c}{Sampling-objective search} \\
    \cmidrule(lr){3-4}\cmidrule(lr){5-6}
    variant & & best $\downarrow$ & feasible / 30 & best $\downarrow$ & feasible / 30 \\
    \midrule
    VAE & $0.398 \pm 0.026$ & $-0.467 \pm 0.116$ & $23.8 \pm 2.3$ & $-0.929 \pm 0.019$ & $29.9 \pm 0.1$ \\
    cVAE & $0.215 \pm 0.021$ & $-0.610 \pm 0.212$ & $28.5 \pm 0.7$ & $-1.021 \pm 0.007$ & $25.5 \pm 0.9$ \\
    \textbf{\method{}} & $\mathbf{0.648 \pm 0.025}$ & $\mathbf{-1.133 \pm 0.006}$ & $\mathbf{28.6 \pm 0.6}$ & $-1.033 \pm 0.000$ & $25.8 \pm 0.5$ \\
    \bottomrule
  \end{tabular}
\end{table*}

The cVAE's novel-objective result varies substantially across seeds
(SE $0.21$): some runs remain on the positive-$x_1$ side of the C
and reach only $f_{\text{opt}}\approx+0.4$.
\method{} achieves mean mid-segment feasibility $0.648$, with
standard error $0.025$ across the $13$ training seeds.

\paragraph{Interpretation.}
C-shape isolates non-convexity within a connected region: connecting
the illustrated endpoints requires a route around the forbidden disc. Shells and the stellarator
experiments test the method in higher dimensions.

\subsection{Shells: setup, interpolation, and optimization}
\label{sec:shells-app}

Shells tests whether regularization helps under strong compression
when distinct mechanisms form disconnected feasible regions.

\begin{figure*}[t]
  \centering
  \includegraphics[width=0.94\linewidth]{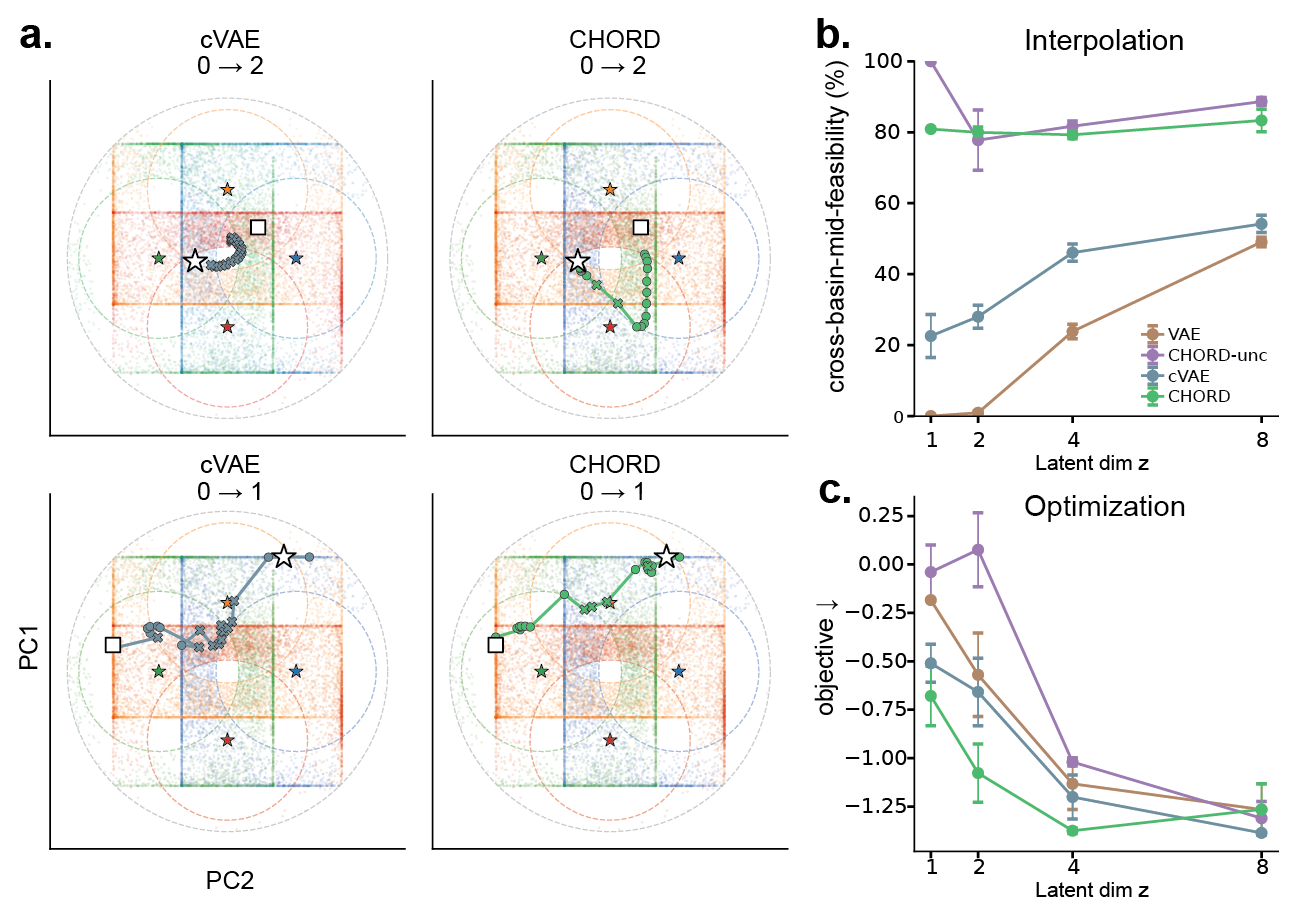}
  \caption{\textbf{Shells benchmark} ($d=16$, $n=3$ seeds).
  \textbf{a.}~Decoded latent chords between two basins for cVAE (left)
  and \method{} (right); circles feasible, crosses not.
  \textbf{b.}~Mid-chord feasibility and \textbf{c.}~held-out objective
  optimization, across latent dimensions.}
  \label{fig:shells-paths}
\end{figure*}

\paragraph{Forward model.}
Four groups of design variables represent alternative mechanisms.
We split a design $x \in \R^{16}$ into $K=4$ disjoint slices
$x_k \in \R^4$. Each slice produces an output through a
shared smooth nonlinearity $\varphi: \R^4 \to \R^2$ plus a
mechanism-specific offset:
\begin{equation*}
\begin{split}
y_k &= \varphi(x_k) + \mu_k,\\
\varphi(v) &= (\tanh(v_0 + v_1),\ \tanh(v_2 + v_3)),
\end{split}
\end{equation*}
with $\mu_k = \rho \cdot (\cos 2\pi k/K, \sin 2\pi k/K)$, $\rho=0.6$:
basin centers on a regular $K$-gon in $y$-space. Soft routing through a
softmax of slice energies determines the dominant mechanism:
\[
\begin{aligned}
\sigma_k &= \mathrm{softmax}_k(\|x_k\|^2 / \tau), \qquad \tau=0.5,\\
y &= \sum_k \sigma_k\, y_k.
\end{aligned}
\]
The simulator output is $y \in \R^2$.

\paragraph{Constraints.}
One mechanism must dominate, and the combined output must lie inside
an outer disc but outside an inner disc centered on the weighted
mechanism offsets:
$$
\begin{aligned}
h_{\text{purity}}(x) &= \tau_p - \max_k \sigma_k \le 0, \\
h_y(x) &= \|y\|_2 - R_y \le 0, \\
h_{\text{shell}}(x) &= r_{\text{in}} - \|y - \textstyle\sum_k \sigma_k \mu_k\|_2 \le 0.
\end{aligned}
$$
Here $\tau_p=0.85$, $R_y=1.4$, and $r_{\text{in}}=0.7$.
Purity forces one mechanism to dominate, separating the four
mechanism regions in $x$-space. A continuous path that changes the
dominant index must pass through a tie, where the largest routing
weight is at most $1/2$, violating the $0.85$ threshold.
The shell also excludes $y=0$, since
$\|\sum_k \sigma_k\mu_k\|\le\rho<r_{\text{in}}$.
A linear metric blend need not pass through the origin, but can cross
this forbidden region. The interpolation diagnostic measures feasibility at sampled interior
points; it cannot certify a continuous feasible connection.

\paragraph{An additional constraint at optimization time.}
To test a requirement absent from training, we introduce
$y_h(x) = \sum_k \sigma_k \|x_k\|^2$ at \emph{optimization time only}.
The constraint $y_h \le y_{h,\max}$ is calibrated to
$\mathrm{q}_{0.7}$ of the CMA-ES feasible distribution
($y_{h,\max}=23.31$). This mirrors the stellarator turbulence-proxy bound $\langle\chi_{\nabla r}\rangle\!\le\!0.85$ ---
a metric never seen by the conditioning vector but applied as a
constraint at optimization time.

The conditioning vector is $c = (y_0, y_1)$ --- the 2-D simulator
output. \textbf{$y_h$ is not in $c$.}

\paragraph{Design choices.}
Table~\ref{tab:shells-knobs} lists the benchmark parameters and their roles.

\begin{table*}[t]
  \centering
  \small
  \caption{Shells parameters and their roles in the benchmark.}
  \label{tab:shells-knobs}
  \begin{tabular}{lll}
    \toprule
    Parameter & Value & Role \\
    \midrule
    $K$ & 4 & provides six distinct pairs of mechanism regions \\
    $d_{\text{per}}$ & 4 & gives a $16$-dimensional design space with four coordinates per mechanism \\
    $\rho$ & 0.6 & basin centres separable but $\|y\| \le R_y$ remains binding \\
    $\tau$ & 0.5 & controls how sharply routing selects the dominant mechanism \\
    $\tau_p$ & 0.85 & requires $85\,\%$ dominance; at $0.7$, cVAE and \method{} perform similarly \\
    $R_y$ & 1.4 & bounds the output norm by $1.4$ \\
    $r_{\text{in}}$ & 0.7 & origin lies inside every basin's forbidden disc since $\|0 - \mu_k\| = \rho < r_{\text{in}}$ \\
    $y_{h,\max}$ & $23.31$ & excludes approximately $30\,\%$ of training designs in the held-out-constraint test \\
    \bottomrule
  \end{tabular}
\end{table*}

\paragraph{Training data.}
For each basin $k$, we draw $20$ feasible starting points by rejection
sampling, increasing the energy of the intended dominant slice.
From each point, CMA-ES runs for $30$ generations with population $8$
and initial step size $\sigma_0=0.30$, minimizing
$f_{\text{sample}}(x)=\|y-\alpha\mu_k\|^2+50\sum_c\mathrm{ReLU}(h_c)$.
We draw $\alpha\sim U[0.4,1.0]$ independently for each start.
Approximately half of the pooled trajectory points are feasible,
yielding about $16$k designs balanced across basins. We then set the
held-out constraint threshold from the feasible sample distribution.

\paragraph{Variants.}
We compare VAE, \method{}-unc, cVAE, and \method{}, each with three
hidden layers of $64$ units. The conditional models receive $y$;
the regularized models include the pathwise feasibility loss.
Four latent dimensions ($z\in\{1,2,4,8\}$) and three training seeds
give $48$ model runs. The sweep ran on \texttt{n2-standard-4} cloud
CPU instances and took approximately eight minutes of wall-clock time.

\paragraph{Diagnostics.}
We retain the archived identifiers in parentheses for comparison
with the original plots.
\begin{description}[leftmargin=1em,style=nextline,itemsep=2pt,topsep=2pt]
  \item[Cross-basin interpolation (T1).]
    We encode validation endpoints from different basins, linearly
    interpolate their latent codes and conditioning over $21$ values
    of $\alpha$, and decode the path. We report mid-segment feasibility
    over $\alpha\in[0.3,0.7]$, averaged over six unordered basin pairs
    with six random endpoint pairs each.
  \item[Local sampling diversity (T2).]
    We encode the training design with the lowest sampling objective
    to obtain $z^\star$. For each $\sigma\in\{0.1,0.3,0.5,1.0\}$,
    we decode $50$ codes $z=z^\star+\sigma\epsilon$, with
    $\epsilon\sim\mathcal{N}(0,I)$ and fixed conditioning.
    We report feasibility and the $5$th--$95$th percentile range of
    $\|y\|^2$ among feasible samples.
  \item[Output-norm optimization (T3).]
    We minimize $\|y\|^2$ with Adam in latent and conditioning space,
    starting from the $30$ training designs with the lowest $\|y\|^2$.
  \item[Novel-objective optimization (T7).]
    We instead minimize $y_0$, which favors the negative-$y_0$
    region associated with basin 2. This changes the search objective;
    $y_0$ remains part of the conditioning vector during training.
  \item[Optimization with a held-out constraint (T5).]
    We minimize the same novel objective while also imposing
    $y_h\le y_{h,\max}$. This constraint is absent from model training.
\end{description}

The output-norm diagnostic checks feasibility separately from
objective value: the shell constraint excludes the origin, so a
small objective value alone does not indicate a successful design.

\paragraph{Results and path variation.}
At the smallest latent dimension, \method{} improves interpolation
and novel-objective search over cVAE (Table~\ref{tab:shells-headline}).
The unconditional regularized model shows why feasibility alone is
insufficient: its decoded paths are nearly constant. Across latent
dimensions, the feasibility gain persists, while the objective
ordering changes at the largest dimension
(Table~\ref{tab:shells-compression}).

\begin{table*}[t]
  \caption{\textbf{Shells results} ($d=16$, $z=1$, $n=3$ seeds, mean$\pm$SE).
  We report mid-segment interpolation feasibility ($\uparrow$) and
  the best feasible value of the novel objective $y_0$ ($\downarrow$),
  with the number of successful optimization starts out of $30$.}
  \label{tab:shells-headline}
  \centering
  \small
  \begin{tabular}{lcc}
    \toprule
    variant & Mid-segment feasibility $\uparrow$ & Best feasible $y_0$ $\downarrow$ (successful starts / 30) \\
    \midrule
    VAE & $0.000$ & $-0.184$\ \ ($4.3 \pm 4.3$) --- collapses \\
    \method{}-unc & $1.000$ & $-0.04\pm0.14$\ \ ($30/30$) --- frozen \\
    cVAE & $0.226 \pm 0.061$ & $-0.510 \pm 0.099$\ \ ($20.0 \pm 5.3$) \\
    \textbf{\method{}} & $\mathbf{0.809 \pm 0.002}$ & $\mathbf{-0.679 \pm 0.154}$\ \ ($\mathbf{28.0 \pm 0.6}$) \\
    \bottomrule
  \end{tabular}
\end{table*}

\begin{figure*}[t]
    \centering
    \includegraphics[width=0.7\linewidth]{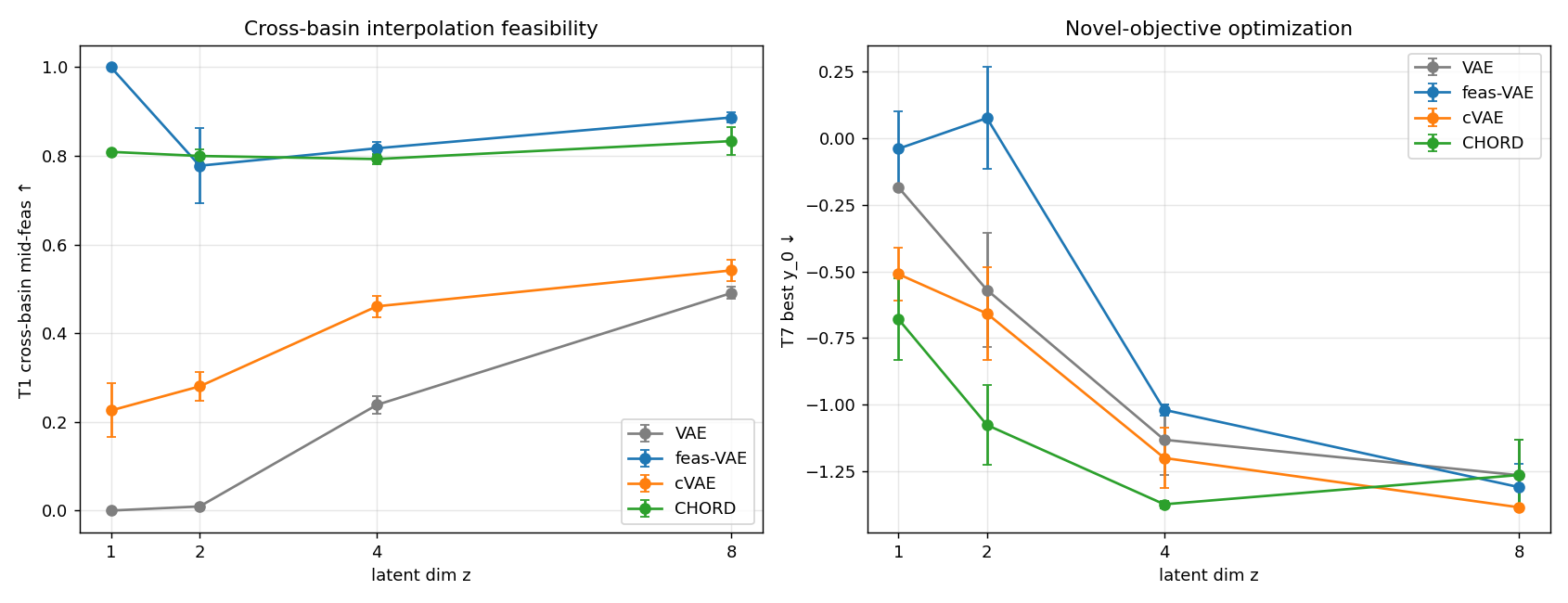}
    \caption{\textbf{Effect of latent dimension on Shells results.}
    The panels show mid-segment interpolation feasibility (T1) and
    best feasible novel-objective value (T7). \method{} improves on
    the cVAE in both measures at $z=1,2,4$; at $z=8$, the cVAE reaches
    a lower objective value.}
    \label{fig:compression-curves-shells}
\end{figure*}

\begin{table*}[t]
  \centering
  \small
  \caption{\method{} and cVAE across latent dimensions.
  \method{} has higher mean interpolation feasibility at every tested
  dimension. Its mean novel-objective value is lower at $z=1,2,4$;
  the cVAE has the lower value at $z=8$. These comparisons do not
  establish why the objective ordering changes.}
  \label{tab:shells-compression}
  \begin{tabular}{lcc}
    \toprule
    $z$ & Feasibility difference (\method{} $-$ cVAE) & Best feasible $y_0$ (\method{} vs cVAE) \\
    \midrule
    1 & $\mathbf{+0.58}$ & $-0.679$ vs $-0.510$ \quad (\method{} $+0.17$) \\
    2 & $\mathbf{+0.52}$ & $-1.077$ vs $-0.658$ \quad (\method{} $+0.42$) \\
    4 & $+0.33$ & $-1.375$ vs $-1.201$ \quad (\method{} $+0.18$) \\
    8 & $+0.29$ & $-1.265$ vs $\mathbf{-1.386}$ \quad (cVAE $+0.12$) \\
    \bottomrule
  \end{tabular}
\end{table*}

\begin{figure*}[t]
    \centering
    \includegraphics[width=0.8\linewidth]{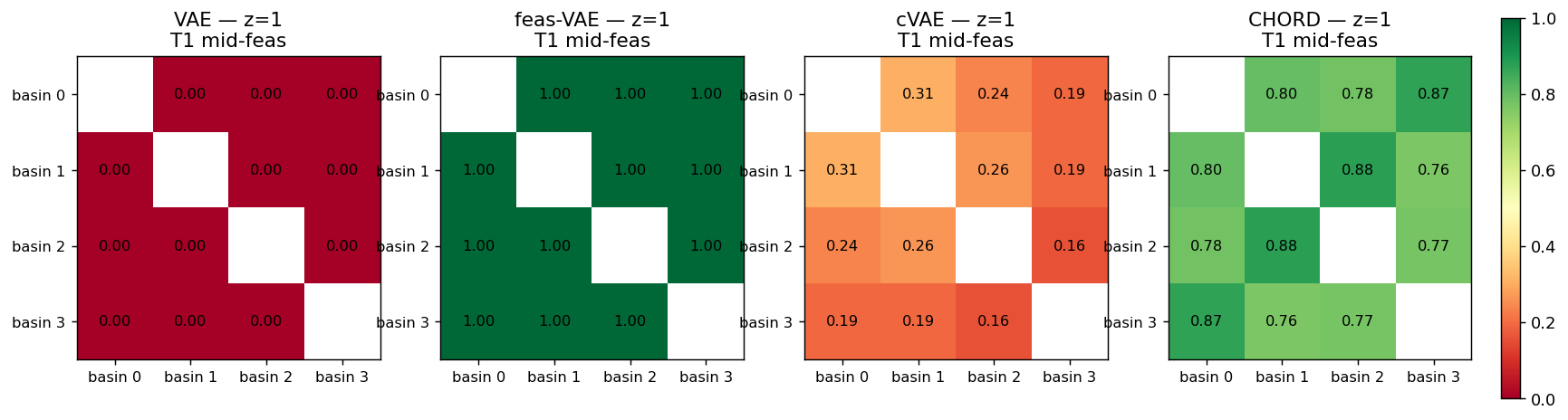}
    \caption{\textbf{Interpolation feasibility by basin pair (T1).}
    Each panel shows mean mid-segment feasibility for one model,
    with cell $(i,j)$ corresponding to endpoints from basins $i$ and
    $j$. The VAE scores $0$ and \method{}-unc scores $1$ throughout;
    the cVAE ranges from $0.16$ to $0.31$, and \method{} from $0.76$
    to $0.88$. The gain over the cVAE occurs across all basin pairs.
    These scores measure sampled interior feasibility, not a
    continuous feasible connection or the extent of design variation.}
    \label{fig:perbasin-t1-heatmap-shells}
\end{figure*}

\begin{figure*}[t]
    \centering
    \includegraphics[width=0.8\linewidth]{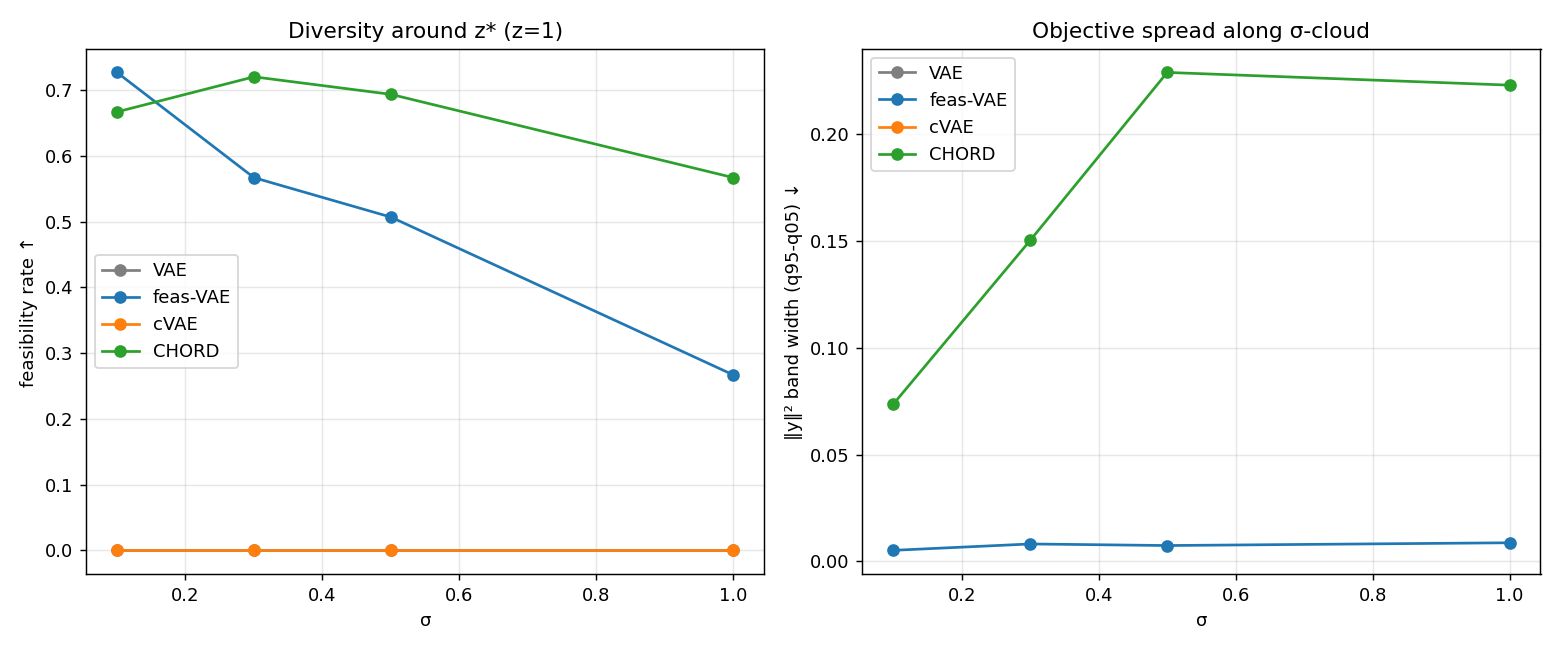}
    \caption{\textbf{Local sampling diversity (T2).} We perturb the
    encoded training design with the lowest sampling objective.
    Left: VAE and cVAE produce no feasible samples at any tested
    noise scale. \method{} retains more than $50\,\%$ feasibility
    through $\sigma=1.0$, whereas \method{}-unc falls below $50\,\%$
    at the largest tested noise scale.
    Right: the $5$th--$95$th percentile range of $\|y\|^2$ among
    feasible samples is nearly zero for \method{}-unc and grows with
    noise scale for \method{}. Thus \method{} combines feasible
    sampling with variation in the evaluated objective.}
    \label{fig:diversity-shells}
\end{figure*}

\begin{figure*}[t]
    \centering
    \includegraphics[width=0.8\linewidth]{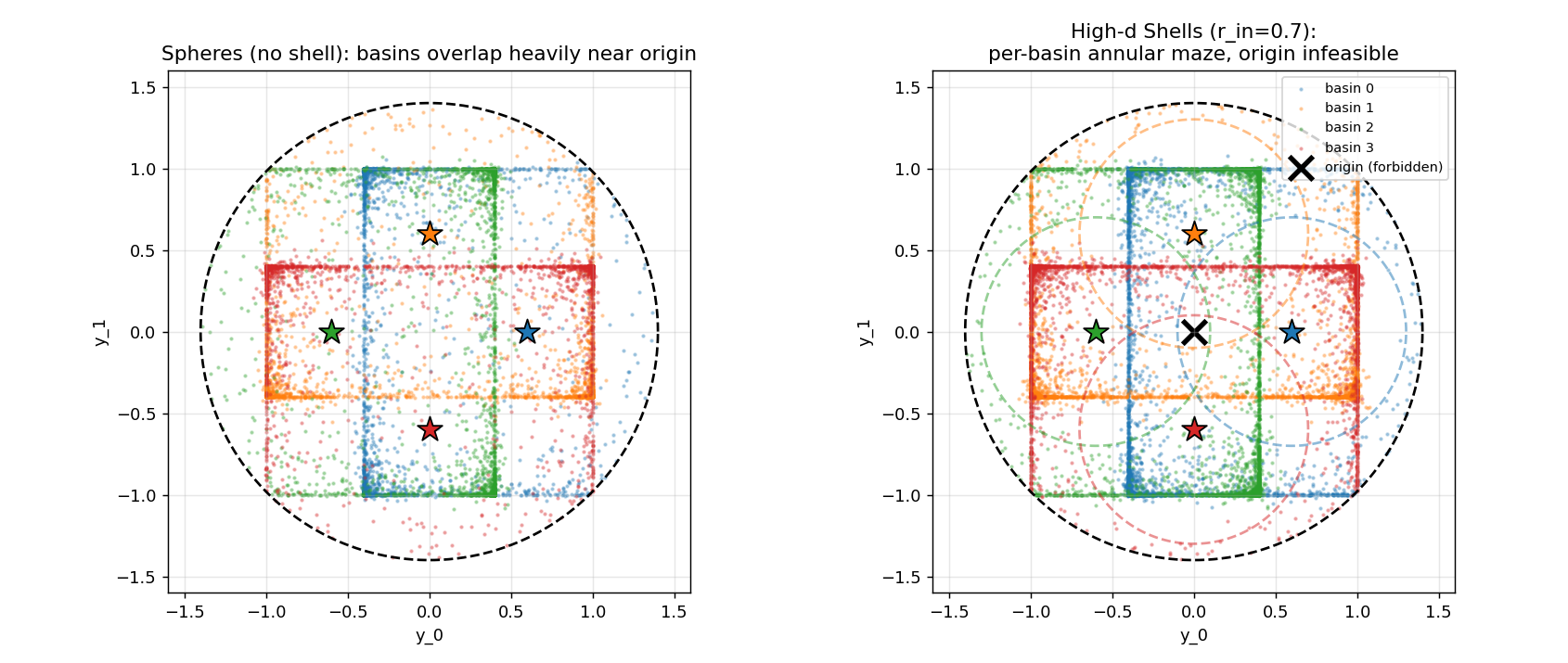}
    \caption{\textbf{Effect of the inner exclusion region.}
    Both output-space plots use purity threshold $\tau_p=0.85$.
    Without the inner exclusion ($r_{\text{in}}=0$, Spheres), the four
    mechanism regions overlap near the origin in output space, so
    similar conditioning values can describe different mechanisms.
    With the exclusion ($r_{\text{in}}=0.7$, Shells), the origin is
    infeasible. This makes the example test interpolation through a
    non-convex feasible output region.}
    \label{fig:speres-vs-shells}
\end{figure*}

\subsection{Stellarator data generation, metrics, and training}
\label{sec:stellarator-setup-app}

\paragraph{Aspect ratio and axis-geometry proxy.}
We define the minor radius by $a=\sqrt{\bar S/\pi}$, where
$\bar S$ is the area enclosed by the plasma boundary in a plane of
constant toroidal angle, averaged over that angle. Thus $a$ is the
radius of a circle with the same average cross-sectional area.
With plasma volume $V$, the corresponding major radius is
$R_{\mathrm{major}}=V/(2\pi\bar S)$ and the aspect ratio is
$A\equiv AR=R_{\mathrm{major}}/a$.
The effective aspect ratio is
$A_{\mathrm{eff}}=\mu_0 I/(2\pi B_0 a)$, where $I$ is the external
poloidal current, $B_0$ the reference magnetic-field strength, and
$a$ the same geometric minor radius defined above. It equals the aspect ratio of an axisymmetric,
flat-axis configuration with the same $I$, $B_0$, and $a$.
We use $A_{\mathrm{eff}}-A$ to distinguish geometry families in this
dataset. Local torsion, in contrast, describes how the magnetic-axis
curve departs from its osculating plane; the proxy is not a torsion
measurement.

\paragraph{Initial guesses and \desc pre-tuning.}
Each augmented-Lagrangian (ALM) campaign with Nevergrad~\cite{bennet2021nevergrad}
starts from a rotating-ellipse boundary parameterised by ellipticity,
axis rotation rate, and field-period count. \vmec sometimes fails to converge on these initial
guesses, leaving QI undefined and preventing the outer optimization
loop from proceeding. We therefore pass each seed
through a brief \desc-based pre-optimization~\cite{dudt2023desc}
and use the resulting boundary to initialize the Nevergrad search.

\paragraph{Sampling protocol.}
Each campaign uses an Augmented Lagrangian outer loop over the
broad-scan-loose constraints with Nevergrad's
\texttt{NgIohTuned} inner optimizer. We bias half of the campaigns toward planar starts and
half toward high-torsion starts to sample both geometry families.
The resulting corpus need not contain equal numbers from each family. The optimization histories contain $5.09$\,M evaluated
boundaries; filtering to broad-scan-loose
feasibility leaves $1.01$\,M. All boundaries are stellarator-symmetric
with three field periods. Generation cost roughly $50$ vCPU-seconds
per evaluated boundary.

\paragraph{Released corpus composition.}
The $100$k-boundary corpus contains $55{,}283$ near-planar-axis and
$44{,}717$ torsion-axis designs. Each row stores $90$ raw Fourier
coefficients, $14$ scalar metrics, the geometry-family label, and
its original training, validation, or test assignment
($80$k/$10$k/$10$k). Preprocessing retains $81$ non-constant
coefficients. The median $A_{\mathrm{eff}}-A$ is $0.27$ in the
near-planar family and $2.42$ in the torsion family.

\paragraph{Metric schema (15 channels).}
\begin{sloppypar}
The frozen surrogate $\hat f$ predicts $15$ scalar metrics from a
boundary in standardized Fourier space, matching the
ConStellaration~\cite{cadena2025constellaration} schema. In the
fixed-aspect-ratio experiments, four inequalities constrain three
metrics: lower and upper bounds on $\iota_{\mathrm{edge}}$, an upper
bound on $\Delta_{\mathrm{edge}}$, and a lower bound on $W_{MHD}$.
The objective, $\log_{10}\mathrm{QI}_{\mathrm{edge}}$, also enters
the conditioning vector but is not a feasibility constraint here.
The six conditioning components are
$y = (A_{\mathrm{eff}}\!-\!AR,\ \iota_{\mathrm{edge}},\
\log_{10}\mathrm{QI}_{\mathrm{edge}},\ \Delta_{\mathrm{edge}},\
W_{MHD},\ \epsilon_{max})$.
The other $11$ channels are $\iota_{\mathrm{axis}}$, $A_{\mathrm{eff}}$,
$AR$, $A_{\mathrm{eff}}-AR$, $\mathrm{max\_elongation}$,
$\mathrm{spectral\_width}$,
normalized global magnetic shear at the edge,
minimum normalized magnetic-gradient scale length,
$\langle\chi_{\nabla r}\rangle_{\rho^2=0.49}$ (the bad-curvature
flux-compression turbulence proxy), its smoothed variant
$\langle\chi_{\nabla r}^{\mathrm{smooth}}\rangle_{\rho^2=0.49}$, and
$\mathrm{magnetic\_mirror\_ratio}_{\mathrm{axis}}$. We report \vmec at
the \texttt{very\_low\_fidelity} preset for the fixed-aspect-ratio
studies. High-fidelity validation is used for the ConStellaration
front-filling results.
\end{sloppypar}

\paragraph{Training hyperparameters.}
All four variants use a residual MLP with three hidden layers of
$512$ units, layer normalization, and the same training loop.
Latent search uses $z=16$; the interpolation study tests
$z\in\{4,8,12,16\}$ (Appendix~\ref{sec:stellarator-interp-app}).
We train for $300$ epochs with batch size $512$, Adam learning rate
$10^{-3}$, and no weight decay. The free-bits threshold is $0.1$
nats per latent dimension. The KL weight follows four $75$-epoch
cycles, rising linearly from $0$ to $0.5$ over the first half of each
cycle and remaining at $0.5$ for the second half. The chord regularizer warms up: $\lambda$
held at $0$ for the first $30$ epochs, ramped linearly to $\lambda=1$
over epochs $30$--$60$, held at $1$ thereafter. The chord penalty
samples $16$ feasible-pair $\alpha$ values per minibatch; the encoder
is detached. Architectures are identical across cVAE-family
($y_{\mathrm{dim}}=6$, or $5$ in the pilot omitting QI) and VAE-family
($y_{\mathrm{dim}}=0$); parameter counts are roughly
$1.16$\,M per VAE-family model and $1.17$\,M per cVAE-family model
(the six conditioning inputs add $6{,}144$ parameters).

\paragraph{Validation reconstruction quality.}
At the reported regularization strength, reconstruction remains
close to the unregularized cVAE. On the $10$k validation split, reconstruction loss ranges from
$0.061$ to $0.066$ in standardized feature space. The unregularized
cVAE lies at the lower end and \method{} with $\lambda=100$ at the
upper end. At the reported setting $\lambda=1$, the gap is below
$1\,\%$. The KL loss converges to approximately $8.0$ nats per design
across variants. PCA projections show similar reconstruction quality
for all four models.

\subsection{Stellarator interpolation: protocol and sensitivity checks}
\label{sec:stellarator-interp-app}

These checks separate interior feasibility from endpoint retention
and test whether the result depends on the scoring grid.

\paragraph{Pair selection and $\alpha$ grid.}
We select $30$ cross-basin pairs from the validation split: one
design from the planar-axis family and one from the torsion-axis
family. The implementation draws sequentially from one NumPy
random generator initialized with the run's pair-selection seed
(default $0$), rather than reseeding for each pair. The saved
design-space endpoints confirm that the same pairs are reused
across training seeds, latent dimensions, variants, and both
$y$-strategies. The analysis manifest records their boundary hashes.

Stored candidates use $\alpha z_a+(1-\alpha)z_b$, with $a$ planar
and $b$ torsion. In the notation of the main text these endpoint
labels are exchanged, so its $\alpha=0$ endpoint is the stored
torsion endpoint.

Per pair we evaluate $21$ uniformly spaced $\alpha$ values in $[0,1]$;
the reported mid-segment statistics average the eight values
$\alpha \in \{0.30,0.35,\ldots,0.65\}$. The original analysis used
floating-point bounds $0.3 \le \alpha \le 0.7$, which excluded the
stored value $0.7000000000000001$. We state the effective grid here
so that the reported figures and tables can be reproduced.
For the comparisons with archived rebuttal baselines we retain this
common eight-point convention. Table~\ref{tab:inclusive-grid}
separately reports the complete main sweep on the symmetric
nine-point grid; the two sets of scores are not pooled.

\paragraph{Self-consistent iteration details.}
We initialize $y_0(\alpha)$ to the linear blend
$(1-\alpha)\,y_a + \alpha\,y_b$ in standardized conditioning space and
apply Eq.~\eqref{eq:self-consistent} for $K=5$ iterations. The
projection $\hat f_y$ takes the surrogate's full $15$-metric output
and selects the six conditioning components. The $\mathrm{clamp}$
operator hard-projects the constrained components
($\iota_{\mathrm{edge}}$, $\Delta_{\mathrm{edge}}$,
$W_{MHD}$) onto their feasible intervals; we use no
buffer margin. The source endpoints come from validation designs
that passed the corpus constraints. Their decoded reconstructions,
including after conditioning updates, are not guaranteed feasible.
Five iterations bring the median per-pair $\Delta y$ below $0.05$ in
standardized conditioning space; subsequent iterations do not measurably
change the decoded boundary.

\paragraph{Comparison across latent dimensions.}
Table~\ref{tab:stellarator-interp-zsweep} reports mid-segment
feasibility (\%), as mean $\pm$ SE across three training seeds.
Each seed contributes the average over $30$ pairs $\times\,8$
mid-$\alpha$ points. Linear-$y$ values for VAE and \method{}-unc
repeat in the self-consistent column because those models have no $y$
to vary.

\begin{table*}[t]
  \centering
  \small
  \caption{Stellarator mid-segment feasibility (\%), mean $\pm$ SE
  across three training seeds at each latent dimension. Boldface
  identifies \method{} with self-consistent conditioning.}
  \label{tab:stellarator-interp-zsweep}
  \setlength{\tabcolsep}{3pt}
  \begin{tabular}{@{}lcccccccc@{}}
    \toprule
    & \multicolumn{4}{c}{Linear $y$ interpolation} & \multicolumn{4}{c}{Self-consistent $y$} \\
    \cmidrule(lr){2-5}\cmidrule(lr){6-9}
    variant & $z{=}4$ & $z{=}8$ & $z{=}12$ & $z{=}16$ & $z{=}4$ & $z{=}8$ & $z{=}12$ & $z{=}16$ \\
    \midrule
    VAE          & $67.5\!\pm\!8.4$ & $71.1\!\pm\!3.3$ & $67.9\!\pm\!3.4$ & $60.1\!\pm\!4.1$ & $67.5\!\pm\!8.4$ & $71.1\!\pm\!3.3$ & $67.9\!\pm\!3.4$ & $60.1\!\pm\!4.1$ \\
    \method{}-unc & $78.8\!\pm\!0.5$ & $66.8\!\pm\!1.1$ & $59.0\!\pm\!1.2$ & $62.6\!\pm\!6.5$ & $78.8\!\pm\!0.5$ & $66.8\!\pm\!1.1$ & $59.0\!\pm\!1.2$ & $62.6\!\pm\!6.5$ \\
    cVAE         & $48.6\!\pm\!0.6$ & $47.8\!\pm\!3.6$ & $50.0\!\pm\!1.0$ & $48.6\!\pm\!3.4$ & $24.7\!\pm\!1.0$ & $32.5\!\pm\!6.1$ & $35.1\!\pm\!0.8$ & $28.8\!\pm\!2.4$ \\
    \method{}    & $30.4\!\pm\!3.2$ & $31.5\!\pm\!1.6$ & $29.3\!\pm\!2.4$ & $29.7\!\pm\!2.7$ & $\mathbf{68.2\!\pm\!4.7}$ & $\mathbf{64.7\!\pm\!3.9}$ & $\mathbf{81.2\!\pm\!4.9}$ & $\mathbf{75.4\!\pm\!3.6}$ \\
    \bottomrule
  \end{tabular}
\end{table*}

\paragraph{Training and inference interact.}
\method{} with self-consistent $y$ has the highest mean feasibility
at $z=12$ and $z=16$, peaking at $81.2\,\%$ at $z=12$.
\method{}-unc leads at $z=4$ and the VAE at $z=8$.
At $z=12$, removing self-consistent inference reduces \method{} from
$81.2$ to $29.3\,\%$; removing the regularizer under the same
self-consistent inference reduces it from $81.2$ to $35.1\,\%$.
The unconditional variants reach $59\text{--}79\,\%$.
For those variants, the regularizer's effect varies with latent
dimension: $+11.3$, $-4.3$, $-8.9$, and $+2.5$ percentage points
at $z=4,8,12,16$, respectively. Per-cell standard errors range from
$0.5$ to $8.4$ percentage points across three seeds.

\paragraph{Sensitivity to the intended symmetric grid.}
Within each latent dimension, the model ordering is unchanged when
the missing interval boundary is included. We rescored all saved candidates in the three-seed main sweep using
the nine samples at $\alpha=0.30,0.35,\ldots,0.70$, including both
interval boundaries with a tolerance of one floating-point step. The reanalysis
reproduces every historical eight-point pair score before changing
the mask. On the nine-point grid, \method{} with self-consistency
reaches $81.9\pm4.3\,\%$ at $z=12$ and $75.4\pm3.5\,\%$ at
$z=16$; design-space interpolation reaches $30.0\,\%$ instead of
$27.9\,\%$. This post-hoc sensitivity check uses
stored simulator outcomes and adds no evaluations. Slerp, pinned-target,
and flow-matching comparisons retain their common historical grid.

\begin{table*}[t]
  \centering\small
  \caption{Nine-point mid-segment feasibility (\%), mean $\pm$ SE
  across three training seeds. Each seed uses the same $30$ pairs.
  This sensitivity table is separate from the historical eight-point comparisons.}
  \label{tab:inclusive-grid}
  \setlength{\tabcolsep}{4pt}
  \begin{tabular}{@{}lcccccccc@{}}
    \toprule
    & \multicolumn{4}{c}{linear conditioning} & \multicolumn{4}{c}{self-consistent conditioning} \\
    \cmidrule(lr){2-5}\cmidrule(lr){6-9}
    model & $z=4$ & $z=8$ & $z=12$ & $z=16$ & $z=4$ & $z=8$ & $z=12$ & $z=16$ \\
    \midrule
    VAE & $69.1\!\pm\!8.5$ & $73.8\!\pm\!3.0$ & $70.7\!\pm\!3.0$ & $63.0\!\pm\!4.2$ & $69.1\!\pm\!8.5$ & $73.8\!\pm\!3.0$ & $70.7\!\pm\!3.0$ & $63.0\!\pm\!4.2$ \\
    \method{}-unc & $81.0\!\pm\!0.3$ & $70.5\!\pm\!1.0$ & $62.5\!\pm\!1.1$ & $65.8\!\pm\!5.9$ & $81.0\!\pm\!0.3$ & $70.5\!\pm\!1.0$ & $62.5\!\pm\!1.1$ & $65.8\!\pm\!5.9$ \\
    cVAE & $48.9\!\pm\!0.6$ & $47.4\!\pm\!3.7$ & $49.1\!\pm\!1.2$ & $47.7\!\pm\!3.3$ & $25.4\!\pm\!0.3$ & $34.4\!\pm\!5.8$ & $36.5\!\pm\!0.1$ & $30.6\!\pm\!1.8$ \\
    \method{} & $34.7\!\pm\!2.6$ & $34.3\!\pm\!2.6$ & $33.8\!\pm\!2.9$ & $34.6\!\pm\!2.6$ & $68.3\!\pm\!4.9$ & $66.2\!\pm\!3.9$ & $81.9\!\pm\!4.3$ & $75.4\!\pm\!3.5$ \\
    \bottomrule
  \end{tabular}
\end{table*}

\paragraph{Endpoint retention and path displacement.}
Interior feasibility does not establish that a path retains its
feasible endpoints. We evaluate the decoded endpoints of the same
saved $21$-point paths; reconstruction and conditioning updates
can lose feasibility (Table~\ref{tab:endpoint-retention}). The design-space baseline also passes only $93.3\,\%$
of endpoint re-evaluations, despite originating from feasible corpus
rows. We therefore distinguish the source labels from feasibility
after preprocessing and simulator re-evaluation. The interior scores
do not establish a feasible path that preserves both endpoints.

We also check whether the paths collapse to a constant design.
At $z=16$, all saved paths have $21$ distinct boundary coefficient
vectors. For \method{} with self-consistency, the median decoded
endpoint separation is $0.876$ times that of the corresponding
design-space endpoints, with minimum $0.234$ across the $90$ paths.
Distances use the Euclidean norm of the $90$ raw Fourier
coefficients. This rules out constant decoded paths in this sample;
it is not a measure of physical distance or a guarantee of geometric
diversity.

\begin{table}[tbp]
  \centering\small
  \caption{Endpoint feasibility at $z=16$ (\%), mean $\pm$ SE across
  three seeds, with $60$ decoded endpoints per seed and method.
  The unconditioned design-space baseline gives $93.3\,\%$.}
  \label{tab:endpoint-retention}
  \setlength{\tabcolsep}{4pt}
  \begin{tabular}{@{}lcc@{}}
    \toprule
    model & linear $y$ & self-consistent $y$ \\
    \midrule
    VAE & $91.7\pm0.0$ & $91.7\pm0.0$ \\
    \method{}-unc & $60.6\pm1.5$ & $60.6\pm1.5$ \\
    cVAE & $74.4\pm4.5$ & $29.4\pm3.4$ \\
    \method{} & $60.6\pm4.3$ & $61.7\pm5.1$ \\
    \bottomrule
  \end{tabular}
\end{table}

\paragraph{Endpoint concentration of intermediate samples.}
We check whether high mid-segment feasibility is accompanied by
samples concentrated near the decoded endpoints. For each path,
we measure distance to the nearer endpoint, normalized by the
separation of the decoded endpoints:
\[
d_{\mathrm{end}}(\alpha)=
\frac{\min\{\|\tilde x(\alpha)-\tilde x(0)\|_2,
\|\tilde x(\alpha)-\tilde x(1)\|_2\}}
{\|\tilde x(1)-\tilde x(0)\|_2}.
\]
Distances use the same raw Fourier coefficients as above. We
average over the historical eight interior points and the same
endpoint pairs, including infeasible candidates to keep this
geometric diagnostic separate from feasibility.

\begin{table*}[t]
  \centering\small
  \caption{Distance of intermediate designs from the nearer decoded
  endpoint, as a fraction of decoded endpoint separation. Values are
  mean $\pm$ SE across three training seeds, each averaging the same
  $30$ pairs and eight interior points. The overall column first
  averages the four latent dimensions within each seed. Conditional
  models use self-consistent inference. This is a post-hoc diagnostic.}
  \label{tab:endpoint-concentration}
  \setlength{\tabcolsep}{4pt}
  \begin{tabular}{@{}lccccc@{}}
    \toprule
    model & $z=4$ & $z=8$ & $z=12$ & $z=16$ & overall \\
    \midrule
    VAE & $0.150 \pm 0.017$ & $0.213 \pm 0.014$ & $0.196 \pm 0.006$ & $0.199 \pm 0.004$ & $0.189 \pm 0.008$ \\
    \method{}-unc & $0.169 \pm 0.003$ & $0.225 \pm 0.004$ & $0.222 \pm 0.013$ & $0.185 \pm 0.005$ & $0.200 \pm 0.004$ \\
    cVAE + self-consistency & $0.357 \pm 0.013$ & $0.334 \pm 0.012$ & $0.347 \pm 0.010$ & $0.331 \pm 0.008$ & $0.342 \pm 0.008$ \\
    \method{} + self-consistency & $0.362 \pm 0.010$ & $0.333 \pm 0.015$ & $0.333 \pm 0.005$ & $0.339 \pm 0.017$ & $0.342 \pm 0.003$ \\
    \bottomrule
  \end{tabular}
\end{table*}

The unconditional models concentrate their intermediate samples
closer to endpoints (Table~\ref{tab:endpoint-concentration}). The
cVAE has a similar distance to \method{}, so this property alone
does not explain the feasibility gain from chord regularization.
The conditional--unconditional ordering persists on the nine-point
grid, when normalizing by source-endpoint separation, and when
considering only simulator-feasible intermediate points. This
measures endpoint concentration in coefficient space; it does not
establish physical diversity or a larger overall path span.

\paragraph{Earlier single-seed campaign and the $\boldsymbol{\lambda}$ sweep.}
An initial pilot used one training seed at $z=16$ with
\texttt{y\_strategy=fixed\_start} (set $y(\alpha)=y_a$ for all $\alpha$)
and $\lambda\in\{0,1,10,100\}$ for the cVAE conditioned on QI.
Mid-segment feasibility declined as $\lambda$ increased from $1$ to
$10$ to $100$. At $\lambda=100$, validation reconstruction loss rose
from $0.062$ to $0.066$ without a feasibility gain. We therefore use
$\lambda=1$ in the fixed-aspect-ratio study. The pilot also tested
conditioning without QI, labelled ``Option B'' in the archive.
That variant was not repeated across training seeds and is absent
from the multi-seed comparison.

\subsection{Stellarator latent optimization: protocol and ablations}
\label{sec:stellarator-opt-app}

We separate the effects of the search coordinates, inference steps,
and model architecture, then account for their computational costs.

\paragraph{Sample drawing and self-consistent iteration.}
We draw $N=10\,000$ codes $z\sim\mathcal{N}(0, I)$ and initial
conditioning vectors $y_0$ with $y_0[\log_{10}\mathrm{QI}]\sim\mathcal{U}(-3.5,-1.5)$,
$y_0[\mathtt{aeff\_minus\_ar}]=3.0$, and the remaining components at
their training-set means in standardized space. We apply
Eq.~\eqref{eq:self-consistent} for $K=30$ iterations with the
clamp projecting $\iota_{\mathrm{edge}}\!\to\![0.745,0.755]$,
$\mathrm{mirror}_{\mathrm{edge}}\!\le\!0.38$,
$\mathrm{vacuum\_well}\!\ge\!0.02$ (i.e.\ feasibility intervals
shrunk by margin $0.02$ on each constrained side); $\log_{10}\mathrm{QI}$
and the other unconstrained components are not clamped. After the
iteration we surrogate-score every decoded design, take the $50$
lowest by predicted $\log_{10}\mathrm{QI}$, and validate them on
real \vmec.

\paragraph{Matched surrogate-assisted search and sampling.}
The main surrogate-assisted CMA-ES comparison returns $0/50$
feasible candidates in the original $81$-D coordinates and $50/50$
in the \method{} latent. These runs share the frozen surrogate and
evaluation budget but use parameterization-specific initializations
and step sizes. Other native-space search methods reach up to
$88\,\%$ feasibility (Fig.~\ref{fig:search-space}).

The failure is consistent with extrapolation beyond the surrogate's
training support: the raw-space candidates end up
$27.6$ standardized units from the nearest training design, where
latent candidates stay near $1.2$, the distance between neighbouring
training rows. At these distances the surrogate scores them at $-4.73$, better
than anything in a $1.01$\,M-row corpus, but \vmec fails to converge
on them. Solver non-convergence does not establish that no
equilibrium exists.

Self-consistent sampling produces a broader spread of candidates
than latent CMA-ES. Across three training seeds, it reaches
$-3.42 \pm 0.17$ in best $\log_{10}\mathrm{QI}$, with all $150$
validated candidates feasible. In the representative comparison,
its $50$ designs span $0.55$ dex in objective value and have mean
pairwise distance $2.57$, compared with $0.001$ dex and $0.20$ for
latent CMA-ES. The corresponding best values are $-3.55$ and $-3.66$.

\paragraph{Inference settings and sensitivity.}
\textbf{Iteration count ($K=30$).} The median $\Delta y$ per sample drops below $0.05$ in
standardized conditioning space by iteration $30$; about half of the
$10\,000$ samples reach $\Delta y < 0.02$.

\textbf{Interior margin ($0.02$).} The margin provides a buffer
against surrogate error. The matched sweep
shows that the no-clamp failures are caused by $W_{MHD}<0$, with
vacuum-well prediction errors of order $10^{-3}$.
The margin is applied to the conditioning vector; its effect on the
decoded design must be checked with \vmec. We swept
seven margins from the same $z$ and $y_0$ draws. At small margins every
infeasible design violated $W_{MHD}$, and $\iota$ never failed. Every
tested margin from $0.01$ to $0.025$ gave $50/50$ feasible designs
(Table~\ref{tab:clamp-margin-sweep}).

\textbf{Initial objective targets ($y_0[\log_{10}\mathrm{QI}]$).}
Sampling in $[-3.5, -1.5]$ biases the iteration toward low QI while
allowing different starting requests. The iteration updates them
using the decoder's predicted metrics. We validate $50$ candidates
to compare the surrogate-assisted methods at a common validation budget.

\paragraph{Comparison with simulator-driven optimization.}
Here the optimizers call \vmec during search.
Table~\ref{tab:stellarator-opt-headline} reports their simulator
budgets separately from the surrogate-assisted comparison above.

\begin{table*}[t]
  \centering
  \small
  \caption{Self-consistent sampling and simulator-driven CMA-ES at
  $z=16$. Self-consistent \method{} uses $50$ \vmec validations;
  the CMA-ES baselines in this table use $200$ \vmec
  evaluations per trial averaged over $60$ trials ($z$-CMA-ES) or
  $120$ trials ($x$-CMA-ES across both pairs).}
  \label{tab:stellarator-opt-headline}
  \setlength{\tabcolsep}{5pt}
  \begin{tabular}{@{}lcccc@{}}
    \toprule
    method & \vmec & best feas $\log_{10}\mathrm{QI}\ \downarrow$ & mean feas $\log_{10}\mathrm{QI}$ & feas.\ rate \\
    \midrule
    \textbf{Self-consistent \method{}} & $\mathbf{50}$ & $\mathbf{-3.55}$ & $\mathbf{-3.15}$ & $\mathbf{50/50}$ \\
    $z$-CMA-ES on \method{}, planar & $200$ & $-3.32$ (best of $60$) & $-2.95$ & $60/60$ trials \\
    $z$-CMA-ES on \method{}, random & $200$ & --- & $-2.82$ & $60/60$ trials \\
    $x$-CMA-ES, planar & $200$ & $-2.00$ (best of $120$) & $-1.83$ & $120/120$ trials \\
    $x$-CMA-ES, random & $200$ & $-2.64$ & $-2.07$ & $90/120$ trials \\
    \midrule
    \emph{Ref.: training min.} ($1.01$\,M pts) & --- & $-3.91$ & --- & --- \\
    \bottomrule
  \end{tabular}
\end{table*}

\begin{table*}[t]
  \centering
  \small
  \caption{Margin sweep for the interior clamp. All runs use the same
  $z$ and $y_0$ draws. Each tested margin from $0.01$ to $0.025$
  gives $50/50$ feasible designs, including the chosen value $0.02$.
  Disabling the clamp gives $3/50$; clamping at the original bounds
  (zero margin) gives $4/50$.}
  \label{tab:clamp-margin-sweep}
  \setlength{\tabcolsep}{5pt}
  \begin{tabular}{@{}lccccccc@{}}
    \toprule
    margin $m$ & no clamp & $0$ & $0.005$ & $0.010$ & $0.015$ & $\mathbf{0.020}$ & $0.025$ \\
    \midrule
    \vmec feasible / 50 & $3$ & $4$ & $22$ & $50$ & $50$ & $\mathbf{50}$ & $50$ \\
    \bottomrule
  \end{tabular}
\end{table*}

\paragraph{Inference ablations ($z=16$, $50$ \vmec each).}
Holding the model fixed isolates the effects of conditioning updates
and interior clamping (Table~\ref{tab:stellarator-opt-mech}).

\begin{table*}[t]
  \centering
  \small
  \caption{Self-consistent latent sampling on \method{} with five
  inference variants. Fixed-$y$ at three QI targets vs.\
  self-consistent inference with and without interior clamping.
  Mean and best are over feasible candidates only.}
  \label{tab:stellarator-opt-mech}
  \begin{tabular}{lccc}
    \toprule
    variant & feas.\ rate & best $\log_{10}\mathrm{QI}\ \downarrow$ & mean $\log_{10}\mathrm{QI}$ \\
    \midrule
    fixed-$y$ at $\log_{10}\mathrm{QI}=-1.5$ (train mean) & $18/50$ & $-2.38$ & $-2.29$ \\
    fixed-$y$ at $\log_{10}\mathrm{QI}=-2.0$ & $2/50$ & $-2.22$ & $-2.21$ \\
    fixed-$y$ at $\log_{10}\mathrm{QI}=-3.0$ & $50/50$ & $-2.04$ & $-1.95$ \\
    self-consistent, no clamp & $3/50$ & $-3.35$ & $-3.33$ \\
    \textbf{self-consistent, clamp $0.02$} & $\mathbf{50/50}$ & $\mathbf{-3.55}$ & $\mathbf{-3.15}$ \\
    \bottomrule
  \end{tabular}
\end{table*}

Fixed conditioning does not reliably turn ambitious requests into
better feasible designs. Only $2/50$ candidates pass \vmec at
$y[\log_{10}\mathrm{QI}]=-2.0$. The most aggressive tested target,
$-3.0$, gives $50/50$ feasible candidates, but their best simulated
value is only $-2.04$.

Self-consistency without clamping reaches low predicted QI, but only
$3/50$ candidates pass \vmec; the other $47$ violate the vacuum-well
bound. Adding interior clamping yields $50/50$ feasible candidates
and best $\log_{10}\mathrm{QI}=-3.55$, with mean QI prediction bias
$+0.06$ dex on those selected candidates.

\paragraph{Model ablation under matched inference ($z=16$, $50$ \vmec each).}
We next hold inference fixed within each model family to isolate
the effect of chord regularization (Table~\ref{tab:stellarator-opt-model}).

\begin{table*}[t]
  \centering
  \small
  \caption{Model comparison under fixed inference schemes. VAE-family models
  have no $y$, so we run unconditional sampling: draw $z$ from the
  prior, decode, surrogate-score, validate top $50$. cVAE-family
  models use the full self-consistent + clamp $0.02$ pipeline.}
  \label{tab:stellarator-opt-model}
  \begin{tabular}{llccc}
    \toprule
    model & inference scheme & feas.\ rate & best $\log_{10}\mathrm{QI}\ \downarrow$ & mean $\log_{10}\mathrm{QI}$ \\
    \midrule
    VAE & unconditional sampling & $43/50$ & $-3.17$ & $-2.61$ \\
    \method{}-unc & unconditional sampling & $42/50$ & $-3.14$ & $-2.65$ \\
    cVAE (no chord reg.) & self-consistent + clamp $0.02$ & $20/50$ & $-2.10$ & $-1.75$ \\
    \textbf{\method{}} & self-consistent + clamp $0.02$ & $\mathbf{50/50}$ & $\mathbf{-3.55}$ & $\mathbf{-3.15}$ \\
    \bottomrule
  \end{tabular}
\end{table*}

Under the same self-consistent inference and clamping, \method{}
improves both feasibility and objective value over the cVAE.
The cVAE yields $20/50$ feasible candidates and \method{} $50/50$,
with best $\log_{10}\mathrm{QI}$ of $-2.10$ and $-3.55$, respectively.

Unconditional VAE and \method{}-unc sampling reach approximately
$85\,\%$ feasibility and best $\log_{10}\mathrm{QI}\approx-3.15$.
Adding the regularizer changes the best value from $-3.17$ to $-3.14$
and the mean from $-2.61$ to $-2.65$.

\paragraph{Surrogate--\vmec calibration.}
We check how objective values change under simulator evaluation.
For each variant we measure mean bias
$\overline{\Delta} = \mathbb{E}[\log_{10}\mathrm{QI}_{\mathrm{\vmec}} - \log_{10}\mathrm{QI}_{\mathrm{surrogate}}]$
on the feasible candidates. Fixed-conditioning variants and
self-consistent sampling with clamping show $|\overline{\Delta}| \le 0.06$.
The no-clamp variant shows $\overline{\Delta} = -0.03$ (slightly
pessimistic at the manifold edge), but only $3/50$ candidates contribute,
so the estimate describes a small selected subset. The Multistart-Adam
result below shows much larger bias ($\sim\!+0.34$ dex).

\paragraph{Multistart Adam from training warm starts.}
To test whether gradient search improves known good designs, we
run surrogate-Adam in $(z, y)$ initialized from the encoder applied
to the $30$ lowest-QI training rows ($500$
Adam iterations, joint $y$ optimization with $|y|\le 4$, $\lambda=10$;
$3$ training seeds, $2$ model pairs, $30$ multistarts each). Best
feasible $\log_{10}\mathrm{QI}=-3.71$ in $30$ \vmec calls; $100\,\%$
feasibility. \textbf{Warm-start comparison:} the $30$ warm-start training
rows already have mean $\log_{10}\mathrm{QI}=-3.83$. Across $180$
trials, Adam's final \vmec value is $0.31$ dex \emph{worse} than the
warm-start row's training value: Adam descends $\sim\!0.18$ dex in
surrogate space, but the surrogate$\to$\vmec bias of $+0.34$ dex erases
that gain. The result largely preserves warm-start quality, without evidence
of improvement over those designs. We report it separately from
Fig.~\ref{fig:search-space} to distinguish this surrogate bias from
the clamped self-consistent results.

\paragraph{Cost breakdown.}
The computational benefit depends on reusing the corpus and models.
Table~\ref{tab:cost-accounting} separates their one-time cost from
the cost of each design query.

\begin{table*}[t]
  \centering
  \small
  \caption{One-time cost against per-query cost. The sampler runs
  $10{,}000$ codes for $30$ iterations, followed by final decoding and
ranking: $3.1\times10^{5}$ decoder and surrogate passes, with no
simulator in the loop. The dataset figure covers the
  campaign that produced the retained corpus; total project spend,
  including exploratory campaigns that were discarded, was about
  $168$k vCPU-hours.}
  \label{tab:cost-accounting}
  \setlength{\tabcolsep}{5pt}
  \begin{tabular}{@{}llll@{}}
    \toprule
    & compute & wall clock & \vmec calls \\
    \midrule
    \emph{one-time} dataset generation & ${\sim}73$k vCPU-h & --- & $5$\,M \\
    \emph{one-time} surrogate training & $1$ GPU & ${\sim}0.5$\,h & $0$ \\
    \emph{one-time} model training & $1$ GPU & ${\sim}0.5$\,h & $0$ \\
    \midrule
    \textbf{per query} self-consistent sampling & $1$ CPU & ${\sim}30$\,s & $0$ \\
    \textbf{per query} validation & cloud \vmec & ${\sim}5$\,min & $\mathbf{50}$ \\
    $z$- or $x$-CMA-ES, per trial & $8$ vCPU & $15$--$25$\,min & $200$ \\
    \bottomrule
  \end{tabular}
\end{table*}

\paragraph{Earlier cold-start CMA-ES + Adam campaign (with turbulence-proxy bound).}
A separate campaign at $z=16$ ran cold-start CMA-ES and Adam with
the additional turbulence-proxy bound
$\langle\chi_{\nabla r}\rangle\!\le\!0.9$
($3$ seeds $\times\,4$ methods $\times\,\{$random, planar$\}\,\times\,20$
optimization seeds per configuration). The bound truncates the
achievable QI floor by $\sim\!1$ dex, so absolute numbers are
shallower; the relative ordering of methods is unchanged. The full
$z$-sweep across $z\in\{4,8,12,16\}$ shows that the
\method{}-vs-cVAE $z$-CMA-ES gap grows with latent dimension
($+0.05$\,dex at $z=4$ to $+0.43$\,dex at $z=16$); we report this
dependence on latent dimension as motivation for using $z=16$
in the main search experiment.

\subsection{Front filling: benchmark, per-gap results, and sensitivity checks}
\label{sec:front-filling-app}

We distinguish the main front-filling evaluation from the smaller
regularizer ablation, then check sensitivity to reference campaigns,
scoring precision, and feasibility tolerance.

\paragraph{Benchmark and metric.}
The MHD-stable track of ConStellaration~\cite{cadena2025constellaration}
scores hypervolume over aspect ratio (minimized) against
$\min L_{\nabla B}\!\times\!\mathrm{NFP}$ (maximized), with reference
point $(-L_{\nabla B}, A) = (1.0, 20.0)$ in minimization space, so the
$L_{\nabla B}$ reference is $-1.0$. Designs are scored at high
fidelity using the benchmark's normalized-violation tolerance of
$0.01$. The nominal thresholds are $|\iota/\mathrm{NFP}| \ge 0.25$,
$\log_{10}\mathrm{QI} \le -3.5$, mirror ratio $\le 0.25$, flux
compression $\le 0.9$ and vacuum well $\ge 0$.
With the tolerance, the corresponding accepted bounds are
$|\iota/\mathrm{NFP}| \ge 0.2475$, $\log_{10}\mathrm{QI} \le -3.465$,
mirror ratio $\le 0.2525$, flux compression $\le 0.909$, and vacuum
well $\ge -0.001$. The last bound uses the benchmark's normalization
scale of $0.1$ for the zero-threshold constraint.
Among the $1{,}200$ evaluated generated designs, $1{,}120$ pass with
this tolerance and $977$ ($81.4\,\%$) pass all nominal bounds strictly.

\paragraph{Metric version for reproduction.}
All metrics here are
computed with \texttt{constellaration}~$0.2.6$. Release $0.3.0$
redefines flux compression in regions of bad curvature from the VMEC
Jacobian to the field-aligned one; since that metric is one of the five
feasibility constraints, the two releases disagree about which designs
are feasible, and reproducing these results requires the $0.2.6$ metric definition.

\paragraph{Anchors and gaps.}
The study retains five campaigns as anchors and six as held-out
comparisons, spanning $A=8$ to $12$. Campaigns at $A=7.2857$,
$7.6786$, and $10.00$ are excluded from these eleven. The last
lies only $0.04$ from the $10.04$ anchor and adds little separation
as a test point. Excluded campaigns remain part of the broader
published benchmark reference, but not the five-anchor baseline.

Our implementation
scores all benchmark-feasible campaign rows at $178.45$, agreeing
with the published $178.4$ to $0.03\,\%$. This broader reference
includes excluded campaigns and extends to $A=7.6786$. Restricting
the comparison to the eleven named anchor and held-out campaigns
gives $176.02$ (Table~\ref{tab:front-reference-scope}).

\begin{table*}[t]
  \centering \small
  \caption{Gap geometry and per-gap results. Mid-gap feasibility is
  the mean over the three blend weights nearest the chord midpoint.
  Corpus-native feasibility
  is the share of existing campaign designs at those aspect ratios
  that already pass the tight thresholds. HF columns report feasible
  survivors / evaluated candidates after selecting each arm's
  low-fidelity in-gap Pareto set. They apply nominal constraints
  without the benchmark allowance. The HF budgets differ by arm, and
  the surviving points are not necessarily non-dominated at HF. These
  counts are separate from the combined front in
  \S\ref{sec:front-filling}.}
  \label{tab:front-filling-gaps}
  \setlength{\tabcolsep}{4.5pt}
  \begin{tabular}{@{}lccccccc@{}}
    \toprule
    & & held-out & \multicolumn{2}{c}{mid-gap feas.\ (\%)}
    & \multicolumn{2}{c}{HF feasible / evaluated} & corpus-native \\
    \cmidrule(lr){4-5}\cmidrule(lr){6-7}
    gap & width & $A$ & $\lambda{=}0$ & $\lambda{=}0.01$
    & $\lambda{=}0$ & $\lambda{=}0.01$ & feas.\ (\%) \\
    \midrule
    g1 & $1.25$ & $8.46$, $8.86$ & $4.6$ & $\mathbf{48.7}$ & $19/38$ & $23/36$ & $6.7$--$8.8$ \\
    g2 & $0.79$ & $9.64$ & $13.6$ & $\mathbf{53.3}$ & $8/23$ & $5/22$ & $16.6$--$19.7$ \\
    g3 & $0.78$ & $10.43$ & $15.3$ & $\mathbf{66.7}$ & $5/22$ & $11/21$ & $13.1$--$24.0$ \\
    g4 & $1.18$ & $11.21$, $11.61$ & $25.1$ & $\mathbf{70.8}$ & $12/26$ & $22/30$ & $16.2$--$18.1$ \\
    \bottomrule
  \end{tabular}
\end{table*}

\begin{table}[tbp]
  \centering\small
  \caption{Front comparisons under the benchmark feasibility
  tolerance. Counts use distinct objective pairs at stored precision.}
  \label{tab:front-reference-scope}
  \setlength{\tabcolsep}{3pt}
  \begin{tabular}{@{}lrr@{}}
    \toprule
    set & front points & hypervolume \\
    \midrule
    five anchors & $16$ & $174.5553$ \\
    anchors + generated & $51$ & $176.3135$ \\
    eleven study campaigns & $33$ & $176.0221$ \\
    broader benchmark & $34$ & $178.4480$ \\
    \bottomrule
  \end{tabular}
\end{table}

\paragraph{Coverage and scoring sensitivity.}
Generated designs add intermediate choices in all four gaps.
In a post-hoc analysis of the stored high-fidelity evaluations, the
combined benchmark front contains $14$ surviving anchor points and
$37$ generated points: $8$, $11$, $5$, and $13$ from gaps g1--g4.
Two original anchor points become dominated. Sorting the front by
aspect ratio, the largest separation between adjacent points falls
from $1.2500$ to $0.5076$.

Exact non-dominated counts depend on numerical resolution. Several
anchor points differ in achieved aspect ratio by about $10^{-9}$.
Rounding both objectives to six decimal places changes the counts
from $16\to51$ to $5\to42$, while leaving the reported hypervolumes
unchanged to four decimal places. Four-decimal rounding also gives
$5\to42$. We retain the benchmark calculation at stored precision
and report coverage separately rather than interpreting the count
as a number of geometrically distinct designs.

Front filling also persists without the benchmark's $1\,\%$
constraint allowance. Applying the nominal bounds to both anchors
and generated designs leaves $85$ anchor rows and $977$ generated
designs feasible. Their combined hypervolume is $172.2466$, up from
$167.7518$ for the strict anchors, and the largest adjacent
aspect-ratio gap falls from $1.25$ to $0.3406$. These are sensitivity
checks on existing evaluations, not additional simulator runs.

\paragraph{Per-gap training.}
Each gap trains its own surrogate and \method{} model on its two
anchor campaigns and augmented designs derived from those campaigns.
We rescale anchor boundaries and re-evaluate them with the simulator;
the resulting training pool includes designs inside the gap.
We exclude held-out campaigns by source-row provenance, including
their rescaled descendants. Filtering only by achieved aspect ratio
would not enforce that separation. Augmentation therefore incurs
additional simulator cost even though it requires no new optimization
campaign.

Saved per-gap configurations use $81$ input coefficients,
$16$ latent dimensions, residual $3\times512$ MLPs, and eight
conditioning metrics: $\iota/\mathrm{NFP}$, $\log_{10}\mathrm{QI}$,
mirror ratio, flux compression, vacuum well, elongation,
magnetic-gradient scale length, and aspect ratio. Training runs for
$300$ epochs at batch size $2048$ and initial learning rate $10^{-3}$,
with a cosine learning-rate schedule, maximum KL weight $0.5$, and
a free-bits threshold of $0.1$ nats per latent dimension. Four
$75$-epoch KL cycles each ramp the weight over their first half and
hold it fixed over their second half. The model-specific chord
weight is $0$ or $0.01$, distinct from the fixed-AR study's $1$.
The penalty emphasizes pairs that span aspect ratio. It adds the
mean violation on generic feasible
pairs to $1.5$ times the mean on pairs separated by at least $0.3$ in
aspect ratio, using up to $32$ pairs per component and uniform
blend weights in $[0,1]$. Each component averages ReLU violations
in the five constrained metrics. The weight ramps from zero over
epochs $30$--$60$, with no inward training margin.

\paragraph{Effect of the regularizer.}
Overall low-fidelity feasibility rises with the regularizer in every gap
($27\!\to\!63$, $45\!\to\!77$, $53\!\to\!85$ and $60\!\to\!86\,\%$),
and the mid-gap gain is between $2.8\times$ and $10.4\times$
(Tab.~\ref{tab:front-filling-gaps}).

\paragraph{Combined front in the separate regularizer ablation.}
This ablation tests whether the regularizer improves the combined
front. The per-gap sweeps together use $11{,}406$ low-fidelity
evaluations and $109$ high-fidelity evaluations for each regularizer
setting. The unregularized arm yields $44/109$ HF-feasible designs;
the regularized arm yields $61/109$. In a post-hoc calculation,
combining each saved HF-feasible set with the nominally feasible
anchors gives hypervolume $172.26$ without regularization and
$172.79$ with it. The largest adjacent aspect-ratio gap decreases
from $0.457$ to $0.383$. We round both objectives to four decimal
places to match the precision of the archived generated points.
These sweeps have equal total evaluation counts, but different
per-gap HF allocations and candidate sets selected by their
low-fidelity Pareto fronts. They support a modest benefit in this
configuration, independently of the separate $1{,}200$-candidate
stratified evaluation; they are not a multi-seed comparison.

\paragraph{Difficulty does not increase monotonically with gap width.}
Gap g4 is easier than the two narrower gaps, g2 and g3.
The widest gap, g1, is the hardest and has the lowest fraction of
feasible source designs (last column of
Tab.~\ref{tab:front-filling-gaps}). Four gaps are insufficient to
identify which properties determine interpolation difficulty.

\paragraph{Surrogate accuracy within each gap.}
On an in-gap holdout, the per-gap surrogates reach physical RMSE
$0.0201$--$0.0210$ against $0.0308$ for a surrogate trained on all
thirteen campaigns, with feasibility accuracy $0.981$--$0.988$ in both
cases. In this comparison, training on the two bracketing campaigns
gives lower in-gap prediction error than training across the broader
aspect-ratio range $7.3$--$12.0$.

\paragraph{Choice of conditioning for front filling.}
We use gap g3, between the anchor campaigns at $A=10.04$ and
$10.82$, as a pilot for choosing the inference scheme. This interval
matches the earlier front-filling experiments. Here, self-consistent
inference produces no feasible designs among $180$ candidates,
although the clamped iteration converges. Improving the surrogate
and varying the tested constraint margins do not recover feasibility.
A midpoint diagnostic shows that self-consistency worsens both QI
and mirror ratio relative to linear conditioning, pushing the latter
above its constraint bound. We therefore use linear conditioning
throughout the front-filling study. Together with the benefit of
self-consistency in the fixed-aspect-ratio experiment
(\S\ref{sec:stellarator-interp}), this pilot motivates choosing the
inference scheme for the task.

\paragraph{Baseline: geometric rescaling.}
Geometric rescaling tests whether a neighbouring solution can reach
a target aspect ratio without a learned generator. At $\Delta A = -0.5$ this
yields $0$ of $9$ feasible boundaries, bound by the vacuum-well
constraint. This small control at one offset does not substitute
for a comparison against retrieving feasible designs from the
augmented training pool. That pool already contains intermediate
aspect ratios. The present experiment measures additions to the
anchor front; it does not establish an advantage over augmented-data
retrieval under a matched high-fidelity evaluation budget.

\paragraph{Scope of the front-filling evaluation.}
High-fidelity evaluation covers $1{,}200$ of $8{,}361$ low-fidelity-feasible
candidates, stratified over ten aspect-ratio bins per gap.
The validated feasible count and attained hypervolume are lower
bounds for the full candidate pool. The feasibility rate is not a
lower bound, and the non-dominated count need not increase when
additional designs dominate existing front points.

The five anchors
already capture $97.8\,\%$ of the broader benchmark hypervolume
($99.2\,\%$ of the eleven-campaign value), so the attainable gain
relative to either reference is small. Aspect-ratio coverage and
hypervolume describe complementary aspects of front filling;
neither establishes geometric diversity.

In the regularizer ablation, the $\lambda{=}0.01$ model
raises mid-gap feasibility in all four gaps. Its screened subsets
yield more HF-feasible survivors in three gaps and fewer in g2.
Per-gap comparisons have different HF budgets; the combined
comparison above has equal totals. The $1{,}200$-candidate benchmark
evaluation covers only the regularized model.

\subsection{Surrogate accuracy, robustness, and ensembling}
\label{sec:surrogate-app}

The surrogate enters the method in two places: it supplies the
feasibility signal in the chord loss at training time, and it ranks
candidates at inference time. It is frozen in both. The decoder can nevertheless exploit errors
in a fixed surrogate, so simulator validation remains necessary.

\paragraph{Accuracy.}
Held-out $z$-scored MSE on the five metrics the pipeline reads is
$0.0007$ for $\iota_{\mathrm{edge}}$, $0.0026$ for vacuum well,
$0.0065$ for $\log_{10}\mathrm{QI}$, $0.0066$ for mirror ratio and
$0.0152$ for $L_{\nabla B}$, on $509$k held-out boundaries with \vmec
ground truth. A deeper $4\times512$ trunk reaches lower training MSE
and essentially the same validation MSE. Increasing depth alone
therefore does not improve held-out accuracy in this comparison.

\paragraph{Swapping the surrogate.}
To test dependence on the ranking surrogate, we retrained three
alternatives on the identical $5.09$M-row corpus: two fresh
initializations and one deeper architecture. Each ranks the same
generated candidate pool.

\begin{table*}[t]
  \centering \small
  \caption{Candidate selection with four independently trained
  ranking surrogates. Overlap counts how many of each top-$50$ match
  the original surrogate's picks.}
  \label{tab:surrogate-swap}
  \setlength{\tabcolsep}{5pt}
  \begin{tabular}{@{}lcccc@{}}
    \toprule
    ranking surrogate & \vmec feasible & best $\log_{10}\mathrm{QI}$
    & surr.\ bias & top-$50$ overlap \\
    \midrule
    original & $50/50$ & $-3.514$ & $-0.005$ & --- \\
    re-init A & $50/50$ & $-3.510$ & $+0.012$ & $12$ \\
    re-init B & $50/50$ & $-3.486$ & $+0.046$ & $12$ \\
    deeper trunk & $49/50$ & $-3.487$ & $+0.032$ & $17$ \\
    \bottomrule
  \end{tabular}
\end{table*}

Across all four ranking surrogates, $199$ of $200$ selected designs
are feasible. The remaining design has
$\iota_{\mathrm{edge}}=0.72457$, below the $0.725$ bound, with a
converged \vmec equilibrium and $\log_{10}\mathrm{QI}=-3.36$.
The best validated QI varies by at most $0.03$ dex, although only
$12$ to $17$ of each surrogate's top-$50$ designs match the original
selection. Near the surrogate's noise floor the ranking is
surrogate-specific, and the top-$50$ \emph{mean} shifts by up to
$0.15$. This test supports robustness of feasibility and best QI to changing
the ranking surrogate; it does not retrain the generative models or
test the whole pipeline with new surrogates. Mean QI varies more:
the spread is $\pm 0.08$ across rankers and $\pm 0.18$ across
generative-model seeds.

\paragraph{Ensembling.}
Averaging the four compatible surrogates improves calibration on the
$509$k held-out rows: feasibility-prediction accuracy rises from
$0.968$ to $0.976$, precision and recall from $0.919$ and $0.921$ to
$0.940$, and $\log_{10}\mathrm{QI}$ RMSE falls from $0.0317$ to
$0.0258$. On the pipeline's own selected candidates the ensemble also
reduces the prediction-error spread from $0.084$ to $0.072$.
All surrogates are slightly optimistic on the selected designs,
predicting $\log_{10}\mathrm{QI}$ values $0.04$ to $0.08$ dex lower
than \vmec. Averaging the predictions reduces their spread but
retains this bias.

\subsection{Controllability: pinning a metric target along the chord}
\label{sec:controllability-app}

Feasibility measures whether a decoded path is usable. It does not
measure whether the path is the one a designer asked for. We test that
separately by pinning one component of the conditioning vector to a
target profile $\mathrm{target}(\alpha)$ at every self-consistent
iteration, leaving the other five components free, and comparing the
simulated metric against the request. The unconditional VAE cannot be
pinned, so we compare its unsteered output with the same targets.
We test three profiles on the same $30$ cross-basin pairs as
\S\ref{sec:stellarator-interp}, using $21$ steps per path and \vmec
evaluation.

\begin{table*}[t]
  \centering \small
  \caption{Controllability under a pinned target. Tracking error is
  the mean absolute deviation of the simulated metric from the
  request, over \vmec-converged steps. Values are mean $\pm$ SE
  across $30$ pairs on one training seed (seed 0).}
  \label{tab:controllability}
  \setlength{\tabcolsep}{5pt}
  \begin{tabular}{@{}llcc@{}}
    \toprule
    profile & method & mid-seg feas.\ (\%) & tracking err.\ (train-$\sigma$) \\
    \midrule
    A: $\iota$ ramp & unconditional VAE (drift) & $67.5 \pm 4.8$ & $\mathbf{1.17 \pm 0.13}$ \\
    & cVAE, pinned & $39.2 \pm 4.7$ & $2.31 \pm 0.19$ \\
    & \method{}, pinned & $\mathbf{81.7 \pm 3.3}$ & $1.45 \pm 0.10$ \\
    \midrule
    B: QI plateau & unconditional VAE (drift) & $67.5 \pm 4.8$ & $\mathbf{0.84 \pm 0.16}$ \\
    & cVAE, pinned & $14.6 \pm 3.4$ & $0.90 \pm 0.07$ \\
    & \method{}, pinned & $\mathbf{77.9 \pm 4.7}$ & $1.06 \pm 0.05$ \\
    \midrule
    C: $\iota \equiv 0.75$ & unconditional VAE (drift) & $67.5 \pm 4.8$ & $1.21 \pm 0.10$ \\
    & cVAE, pinned & $40.8 \pm 4.8$ & $2.55 \pm 0.22$ \\
    & \method{}, pinned & $\mathbf{79.2 \pm 3.4}$ & $1.23 \pm 0.07$ \\
    \bottomrule
  \end{tabular}
\end{table*}

\paragraph{Feasibility and target-tracking accuracy.}
\method{} retains $77.9$ to $81.7\,\%$ feasibility under pinning on
the seed-0 checkpoint. This is descriptively similar to the unpinned
three-seed result, but is not a matched multi-seed test of the cost
of pinning. The pinned cVAE reaches $14.6$ to $40.8\,\%$.
Under the pre-specified mean absolute tracking-error criterion,
\method{} does not improve on the unconditional VAE on any profile. A linear blend of endpoint metrics is not necessarily
the metric trajectory of an unsteered decoded chord.

\paragraph{Post-hoc analysis of QI loss.}
Absolute tracking error penalizes improvements as well as
deteriorations relative to the requested QI. We chose this one-sided measure after examining
the pre-specified absolute-error result; it is a post-hoc analysis.
For profile B, we examine signed error and one-sided QI degradation,
$\mathrm{mean}\,\max(0,\log_{10}\mathrm{QI}_{\mathrm{simulated}}
-\log_{10}\mathrm{QI}_{\mathrm{target}})$.
The unconditional VAE has signed error $+0.203\pm0.043$ dex and
one-sided degradation $0.224\pm0.044$ dex. The cVAE's degradation is
$0.188\pm0.024$ dex. \method{} has signed error $-0.106\pm0.029$
dex and degradation $0.101\pm0.017$ dex. Its paired difference from
the VAE is $-0.123\pm0.044$ dex, approximately $2.8$ standard errors.
Thus \method{} improves on the target on average while still losing
QI at some steps.

\subsection{Additional baselines}
\label{sec:baselines-app}

\paragraph{Interpolation paths.}
Slerp follows a spherical path between latent codes rather than a
straight chord. We compare both paths using the same seed-0
checkpoints, $30$ cross-basin validation pairs, $21$ interpolation
steps, and \vmec evaluation procedure. Table~\ref{tab:path-baselines} reports the
eight-point interior statistic used in the main experiment.
Unlike the main multi-seed table, its uncertainty is the standard
error across endpoint pairs. Slerp improves the unconditional VAE by
$4.2$ percentage points but changes \method{} by $-1.3$ points.
\method{} with slerp exceeds VAE+slerp by about $9$ percentage points on this
checkpoint. The evidence supports a useful conditional model here;
it neither establishes a multi-seed advantage over VAE+slerp nor
shows that slerp improves \method{}.

\begin{table*}[t]
  \centering \small
  \caption{Matched path comparison at $z=16$, training seed 0.
  Mean $\pm$ SE across the same $30$ endpoint pairs; every decoded
  boundary is evaluated with \vmec. The linear-path rows use the
  saved seed-0 results, rather than the three-seed averages.}
  \label{tab:path-baselines}
  \setlength{\tabcolsep}{5pt}
  \begin{tabular}{@{}llcc@{}}
    \toprule
    model & conditioning & linear path (\%) & slerp (\%) \\
    \midrule
    unconditional VAE & none & $67.5\pm4.8$ & $71.7\pm5.0$ \\
    \method{} & self-consistent, $K=5$ & $81.7\pm2.7$ & $80.4\pm3.5$ \\
    \bottomrule
  \end{tabular}
\end{table*}

\paragraph{Flow-matching backbone.}
To test the method beyond autoencoders, we train a conditional
flow-matching model~\cite{lipman2023flowmatching} on the same $100$k-boundary dataset and split, with the same six
conditioning metrics. The velocity field uses a three-layer,
$512$-wide residual MLP ($647$k parameters, versus $582$k in the
cVAE decoder) and trains for $150$ epochs. It transports Gaussian
noise to boundary coefficients through an ordinary differential
equation (ODE). To interpolate, we integrate backward from each
endpoint to obtain its noise code, interpolate those codes, and
integrate forward with interpolated conditioning. The noise space
has $81$ dimensions, so this is a comparison of generator families
at similar decoder capacity, not at matched bottleneck dimension.
Evaluation uses $100$-step Heun integration.

We also train a chord-regularized flow: invert paired endpoints
without gradients, interpolate their codes and conditioning, decode
with a differentiable ODE solve, and penalize surrogate-predicted
constraint violation. We choose $\lambda=50$ from a seed-0 sweep of
$\{50,150,400\}$, then keep it fixed across three training seeds.
The inference iteration count and clamp margins are inherited from
the cVAE experiment rather than tuned for the flow.

\begin{table}[H]
  \centering \small
  \caption{All flow-matching interpolation ablations. Mean $\pm$ SE
  across three training seeds, each evaluated on the same $30$
  cross-basin pairs with the eight-point interior statistic.}
  \label{tab:flow-baselines}
  \setlength{\tabcolsep}{3pt}
  \begin{tabular}{@{}llcc@{}}
    \toprule
    path & conditioning & \shortstack{base flow\\(\%)} & \shortstack{chord-regularized\\flow (\%)} \\
    \midrule
    linear & linear & $59.9\pm4.4$ & $34.6\pm13.6$ \\
    slerp & linear & $50.8\pm5.1$ & $32.1\pm10.3$ \\
    slerp & self-consistent & $14.0\pm2.1$ & $32.8\pm6.7$ \\
    \bottomrule
  \end{tabular}
\end{table}

Every tested regularized variant remains below the base flow with
linear interpolation (Table~\ref{tab:flow-baselines}). Under
self-consistent inference, the chord penalty raises feasibility
from $14.0$ to $32.8\,\%$, but this does not establish a performance
benefit over the base flow. A broader tuning study would be needed
to compare the best attainable performance of the model families.

\paragraph{Cycle-consistency penalty for latent search.}
Notin et al.~\cite{notin2021decoderuncertainty} guide latent search
using epistemic decoder uncertainty: uncertainty about the learned
decoder parameters. They estimate mutual information between those
parameters and generated outputs using importance sampling, with
Monte Carlo dropout to sample an approximate parameter posterior.
Our \method{} checkpoint was trained without dropout. The diagnostic
we tested instead inserts dropout at inference and measures variation
in decoded coefficients. This is a perturbation statistic, not their
posterior-based mutual-information estimator.

To keep the trained representation fixed, we use a deterministic
latent cycle error as the search penalty:
\[
r_{\mathrm{cyc}}(z,y)=
\bigl\|z-E_\phi(D_\theta(z,y),y)\bigr\|_2,
\]
where $E_\phi(\cdot,y)$ denotes the encoder mean with conditioning
shown explicitly. This measures agreement between one fixed encoder
and decoder, rather than uncertainty across decoder parameters.
The criterion is related to the latent consistency studied by
Boyar and Takeuchi~\cite{boyar2024lsbo}; we apply it during search
without their full LCA-LSBO procedure.

We selected this proxy because it correlates with distance to nearby
training samples: Spearman $\rho=0.64$, versus $0.03$ for the
inference-time dropout statistic. The distance is measured in
standardized boundary coefficients against a reference subset of
training designs. This comparison assesses distance sensitivity,
not calibration of probabilistic uncertainty.

We add $\beta r_{\mathrm{cyc}}$ to the same surrogate-based CMA-ES
score as the latent baseline, retaining its feasibility penalties
and conditioning updates. A reduced-budget sweep over
$\beta\in\{0,0.1,0.3,1\}$ selects $\beta=0.3$ for comparison with
$\beta=0$ at $300$k surrogate evaluations per arm. Both use the same
frozen $z=16$, seed-0 \method{} model and a validation budget of up
to $50$ distinct candidates. We rank surrogate-feasible candidates
by the penalized score and require a minimum separation of $0.1$
in standardized coefficient space. The penalty arm yields only nine
such candidates, so only nine are validated.

\begin{table}[H]
  \centering \small
  \caption{Latent-search comparisons. The penalty arm uses $300$k
  surrogate evaluations and validates its nine distinct candidates;
  the other arms validate $50$ designs. Self-consistent sampling
  additionally uses $10$k surrogate calls for final ranking.}
  \label{tab:notin}
  \setlength{\tabcolsep}{3pt}
  \begin{tabular}{@{}lcc@{}}
    \toprule
    method & feasible & best $\log_{10}\mathrm{QI}$ \\
    \midrule
    cycle penalty ($\beta{=}0.3$) & $5/9$ & $-3.351$ \\
    latent CMA-ES ($\beta{=}0$) & $50/50$ & $\mathbf{-3.661}$ \\
    self-consistent sampling & $50/50$ & $-3.551$ \\
    training-corpus retrieval & $50/50$ & $-3.893$ \\
    \bottomrule
  \end{tabular}
\end{table}

On this checkpoint, the additional cycle penalty reduces candidate
yield and simulator feasibility while worsening the best QI
(Table~\ref{tab:notin}). Latent CMA-ES and self-consistent sampling
already find feasible, better-QI designs without that penalty.
The comparison therefore supports omitting this cycle penalty in
the tested search; it does not evaluate Bayesian decoder uncertainty.

\paragraph{Connectivity of the sampled design corpus.}
We ask whether stored feasible designs already form routes between
the sampled geometry families. A $k$-nearest-neighbour graph over $80$k training designs connects
$0$ of $30$ cross-basin pairs at $k \le 20$ ($135$ and $106$
components). At $k=50$ only $16$ of $30$ connect, with path lengths
about $3.7\times$ the Euclidean distance. The graph vertices are feasible stored designs, but feasibility
between adjacent vertices has not been established. Graph paths
exist for only $16/30$ pairs at $k=50$ and provide no validated
continuous parameterization. This demonstrates sparse connectivity
in the sampled corpus, not topological disconnection of the full
feasible set.

\end{document}

%% file: figures/method.tex
\begin{tikzpicture}[
    >={Stealth[length=2.0mm,width=1.6mm]},
    font=\footnotesize,
    every node/.style={inner sep=1pt, outer sep=0pt}
]

\node[anchor=south west, inner sep=0pt] at (0,0)
    {\includegraphics[width=16cm,height=9cm]{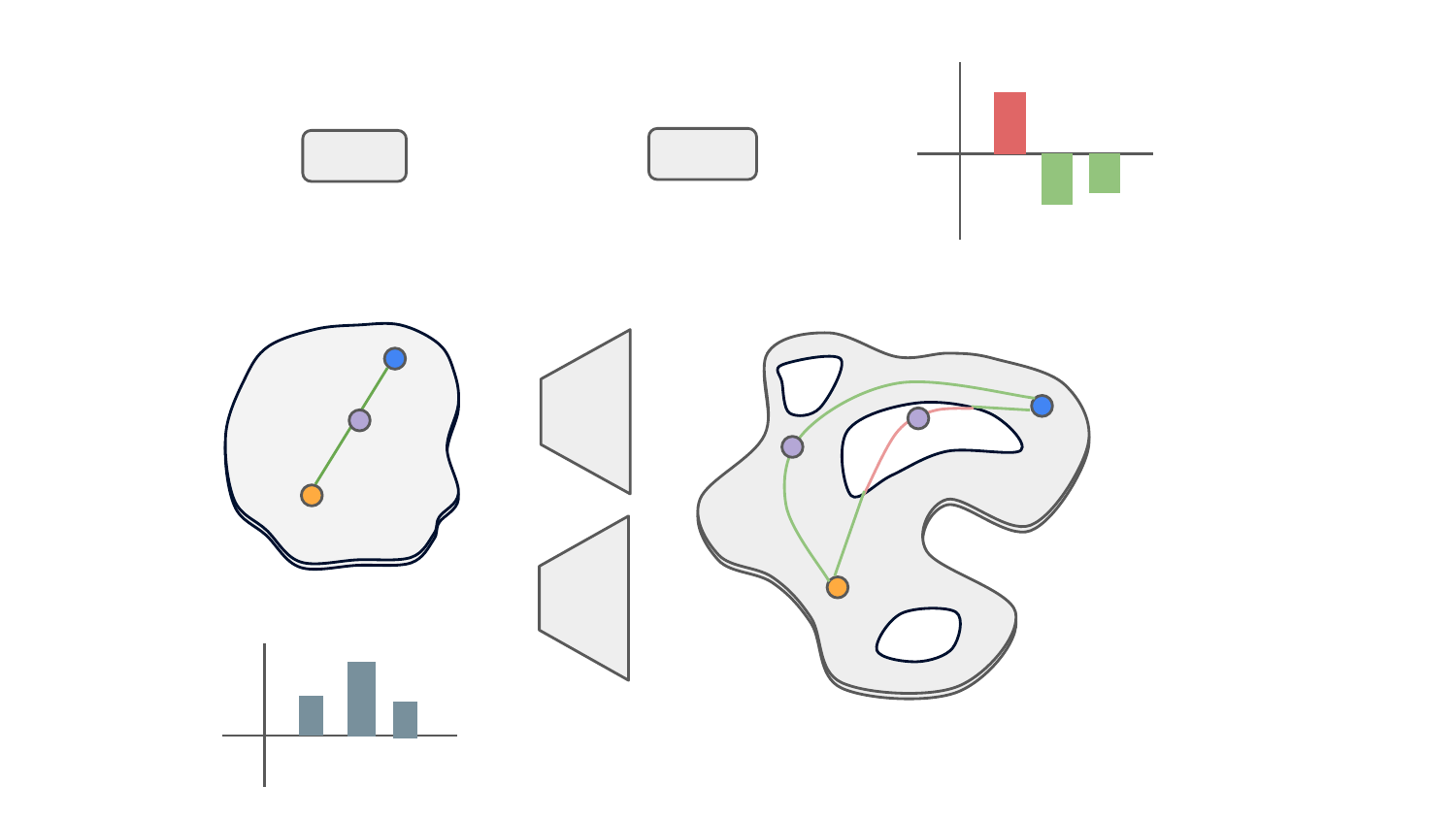}};


\node[font=\bfseries\large, anchor=west] at \sv{75}{45} {a.};

\node[font=\small] at \sv{145}{30}  {Input};
\node[font=\small] at \sv{145}{52}  {designs};
\node[font=\small] at \sv{235}{40}  {Simulator};
\node[font=\small] at \sv{365}{40}  {Metrics};
\node[font=\small] at \sv{540}{40}  {Constraints};

\node[font=\normalsize] at \sv{145}{105} {$x$};
\draw[->, thick, color=labelgray] \sv{160}{105} -- \sv{198}{105};

\node[font=\small\itshape] at \sv{235}{105} {$f(x)$};
\node[font=\small\itshape] at \sv{465}{105} {$h(y)$};

\draw[color=labelgray, thick] \sv{272}{105} .. controls \sv{305}{105} and \sv{325}{75}  .. \sv{350}{75};
\draw[color=labelgray, thick] \sv{272}{105} .. controls \sv{305}{105} and \sv{325}{105} .. \sv{350}{105};
\draw[color=labelgray, thick] \sv{272}{105} .. controls \sv{305}{105} and \sv{325}{170} .. \sv{350}{170};

\node[font=\small\itshape] at \sv{365}{75}  {$y_1$};
\node[font=\small\itshape] at \sv{365}{105} {$y_2$};
\node[font=\normalsize]    at \sv{365}{135} {$\vdots$};
\node[font=\small\itshape] at \sv{365}{170} {$y_D$};

\draw[color=labelgray, thick] \sv{380}{75}  .. controls \sv{410}{75}  and \sv{420}{105} .. \sv{428}{105};
\draw[color=labelgray, thick] \sv{380}{105} -- \sv{428}{105};
\draw[color=labelgray, thick] \sv{380}{170} .. controls \sv{410}{170} and \sv{420}{105} .. \sv{428}{105};

\draw[color=labelgray, thick] \sv{500}{105} .. controls \sv{530}{105} and \sv{545}{75}  .. \sv{555}{75};
\draw[color=labelgray, thick] \sv{500}{105} .. controls \sv{530}{105} and \sv{545}{105} .. \sv{555}{105};
\draw[color=labelgray, thick] \sv{500}{105} .. controls \sv{530}{105} and \sv{545}{170} .. \sv{555}{170};

\node[font=\small\itshape] at \sv{580}{75}  {$h_1$};
\node[font=\small\itshape] at \sv{580}{105} {$h_2$};
\node[font=\normalsize]    at \sv{580}{135} {$\vdots$};
\node[font=\small\itshape] at \sv{580}{170} {$h_K$};

\node[font=\small\itshape] at \sv{615}{30}    {$h_k$};
\node[font=\footnotesize\itshape] at \sv{667}{115} {$h_1$};
\node[font=\footnotesize\itshape] at \sv{697}{150} {$h_2$};
\node[font=\footnotesize\itshape] at \sv{728}{145} {$h_3$};

\node[font=\small, color=violationred, anchor=west] at \sv{687}{60}  {$h_k\!>\!0$: violation};
\node[font=\small\itshape, color=feasiblegreen, anchor=west] at \sv{687}{180} {$h_k\!\le\!0$: feasible};


\node[font=\bfseries\large, anchor=west] at \sv{75}{215} {b.};

\node[font=\small] at \sv{225}{198} {Latent space};
\node[font=\small] at \sv{565}{198} {Parameter space};

\node[font=\small\itshape] at \sv{272}{225} {$z_i$};
\node[font=\small\itshape, anchor=east]  at \sv{220}{280} {$z(\alpha)$};
\node[font=\small\itshape] at \sv{195}{348} {$z_j$};

\node[font=\scriptsize, color=chordgreen, align=center] (lcord) at \sv{295}{378} {Linear chord};
\draw[->, color=chordgreen, thin]
    \sv{280}{370} .. controls \sv{260}{350} and \sv{240}{330} .. \sv{225}{305};

\node[font=\normalsize\itshape] at \sv{386}{272} {$E_\phi$};
\node[font=\normalsize\itshape] at \sv{384}{395} {$D_\theta$};

\draw[->, color=labelgray, thick]
    \sv{527}{290} .. controls \sv{490}{260} and \sv{460}{255} .. \sv{418}{262};

\draw[->, color=labelgray, thick]
    \sv{356}{262} .. controls \sv{310}{262} and \sv{275}{272} .. \sv{248}{277};

\draw[->, color=labelgray, thick]
    \sv{248}{282} .. controls \sv{300}{310} and \sv{340}{340} .. \sv{358}{370};

\draw[->, color=labelgray, thick]
    \sv{414}{378} .. controls \sv{460}{382} and \sv{500}{320} .. \sv{517}{300};

\node[font=\footnotesize, anchor=west] at \sv{145}{398}  {Conditioning};
\node[font=\footnotesize, anchor=west] at \sv{145}{418}  {metrics at $\alpha$};
\node[font=\small\itshape] at \sv{152}{435} {$y_d$};
\node[font=\footnotesize\itshape] at \sv{205}{510} {$y_1$};
\node[font=\footnotesize\itshape] at \sv{238}{510} {$y_2$};
\node[font=\footnotesize\itshape] at \sv{267}{510} {$y_3$};
\draw[->, color=labelgray, thick]
    \sv{295}{475} .. controls \sv{340}{475} and \sv{370}{465} .. \sv{378}{452};

\node[font=\small\itshape, anchor=north] at \sv{687}{282} {$x_i$};
\node[font=\small\itshape, anchor=north] at \sv{522}{310} {$x(\alpha)$};
\node[font=\small\itshape, anchor=west] at \sv{563}{395} {$x_j$};

\node[font=\tiny\itshape, color=labelgray] at \sv{615}{395} {feasible};

\node[font=\small\bfseries, anchor=west] at \sv{735}{218} {\texttt{CHORD}};
\draw[->, thick, color=labelgray]
    \sv{735}{226} .. controls \sv{715}{245} and \sv{685}{255} .. \sv{665}{260};

\node[font=\footnotesize, anchor=west] at \sv{740}{370} {cVAE};
\draw[->, thick, color=labelgray]
    \sv{738}{367} .. controls \sv{700}{350} and \sv{640}{320} .. \sv{585}{290};

\draw[color=labelgray, thick, ->] \sv{505}{475} -- \sv{485}{490}; 
\draw[color=labelgray, thick, ->] \sv{505}{475} -- \sv{505}{435}; 
\draw[color=labelgray, thick, ->] \sv{505}{475} -- \sv{545}{475}; 



\node[anchor=west, font=\footnotesize] at \sv{125}{555}
{$\mathcal{L}_{\text{chord}} = \mathbb{E}_{(i,j),\alpha}\!\left[\,\sum_{k=1}^{K} w_k\!\left[h_k\!\left(\hat{f}\!\left(D_\theta(z_{ij}(\alpha),\, y_{ij}(\alpha))\right)\right)\right]_{+}\right]$};

\node[anchor=west, font=\footnotesize\bfseries] at \sv{640}{530} {Minimize};
\node[anchor=west, font=\footnotesize] at \sv{605}{555}
{$\mathcal{L} = \mathcal{L}_{\text{cVAE}} + \lambda_{\text{chord}}\,\mathcal{L}_{\text{chord}}$};

\end{tikzpicture}